\documentclass[sigconf]{acmart}

\usepackage{microtype}
\usepackage{graphicx}
\usepackage{subfigure}
\usepackage{multirow}
\usepackage{natbib}
\usepackage{hyperref}
\usepackage{bm}
\usepackage{amsmath}
\usepackage{enumitem}
\DeclareMathOperator*{\minimize}{minimize}
\DeclareMathOperator*{\maximize}{maximize}

\usepackage{amsmath}
  
\usepackage{amssymb}

\usepackage{mathtools}
\usepackage{amsthm}

\usepackage[capitalize,noabbrev]{cleveref}

\theoremstyle{plain}

\theoremstyle{definition}

\theoremstyle{remark}

\acmConference[KDD '23]{The 29th ACM SIGKDD Conference on Knowledge Discovery and Data Mining}{August 6--10, 2023}{Long Beach, California}
\copyrightyear{2023} 
\acmYear{2023} 
\setcopyright{acmlicensed}\acmConference[KDD '23]{Proceedings of the 29th ACM SIGKDD Conference on Knowledge Discovery and Data Mining}{August 6--10, 2023}{Long Beach, CA, USA}
\acmBooktitle{Proceedings of the 29th ACM SIGKDD Conference on Knowledge Discovery and Data Mining (KDD '23), August 6--10, 2023, Long Beach, CA, USA}
\acmPrice{15.00}
\acmDOI{10.1145/3580305.3599337}
\acmISBN{979-8-4007-0103-0/23/08}

\begin{document}

% The \icmltitle you define below is probably too long as a header.
% Therefore, a short form for the running title is supplied here:
\title{ExplainableFold: Understanding AlphaFold Prediction with Explainable AI}

\author{Juntao Tan}
\affiliation{%
  \institution{Rutgers University, New Brunswick, NJ, US}
  \country{}
}
\email{juntao.tan@rutgers.edu}

\author{Yongfeng Zhang}
\affiliation{%
  \institution{Rutgers University, New Brunswick, NJ, US}
  \country{}
}
\email{yongfeng.zhang@rutgers.edu}

\renewcommand{\shortauthors}{Juntao Tan and Yongfeng Zhang}

\begin{abstract}

This paper presents ExplainableFold ($x$Fold), which is an Explainable AI framework for protein structure prediction. Despite the success of AI-based methods such as AlphaFold ($\alpha$Fold) in this field, the underlying reasons for their predictions remain unclear due to the black-box nature of deep learning models. To address this, we propose a counterfactual learning framework inspired by biological principles to generate counterfactual explanations for protein structure prediction, enabling a dry-lab experimentation approach. Our experimental results demonstrate the ability of ExplainableFold to generate high-quality explanations for AlphaFold's predictions, providing near-experimental understanding of the effects of amino acids on 3D protein structure. This framework has the potential to facilitate a deeper understanding of protein structures. Source code and data of the ExplainableFold project are available at \url{https://github.com/rutgerswiselab/ExplainableFold}.

\end{abstract}


\maketitle

\section{Introduction}
\label{sec:introduction}

The protein folding problem studies how a protein's amino acid sequence determines its tertiary structure. It is crucial to biochemical research because a protein's structure influences its interaction with other molecules and thus its function. Current machine learning models have gained increasing success on 3D structure prediction \cite{alquraishi2021machine, torrisi2020deep}. Among them, AlphaFold \cite{jumper2021highly} provides near-experimental accuracy on structure prediction, which is considered an important achievement in recent years. Nevertheless, one of the significant challenges with AlphaFold, as well as other deep learning models, is that they cannot provide explanations for their predictions. Essentially, the \textit{why} question still remains largely unsolved: the model gives limited understanding of why the proteins are folded into the structures they are, which hinders the model's ability to provide deeper insights for human scientists.

However, explainability is a critically perspective in AI for Science research (Explainable AI for Science) \cite{li2022from}, since science is not only about understanding the ``how'', but also, and perhaps more importantly, the ``why''. Specifically, in protein structure prediction research, it is crucial to understand the mechanism of protein folding from both AI and scientific perspectives. From the AI perspective, explainability has long been an important consideration. State-of-the-art protein structure prediction models leverage complex deep and large neural networks, which makes it difficult to explain their predictions or debug the trained model for further improvement. From the scientific perspective, scientists' eagerness to conquer knowledge is not satisfied with just knowing the prediction results, but also knowing the \textit{why} behind the results \cite{li2022from}. In particular, structural biologists not only care about the structure of proteins, but also need to know the underlying relationship between protein primary sequences and tertiary structures \cite{dill2008protein, dill2012protein}.

\begin{figure*}[t]
    \centering
    \includegraphics[width=.9\linewidth]{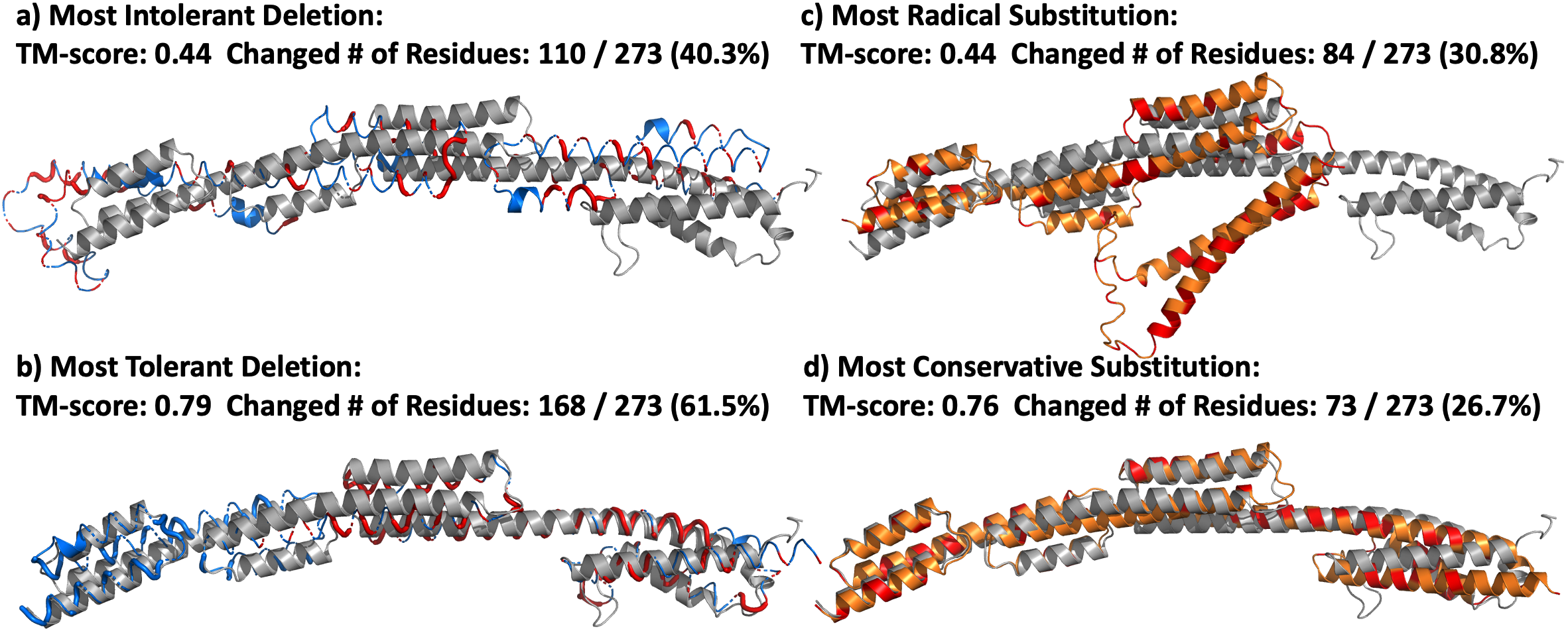}
    \vspace{-10pt}
    \caption{The original protein is colored gray; structures following amino acid deletion and substitution are blue and orange, respectively, with red indicating the altered residues. (a) Some amino acids play crucial roles in protein folding. By removing the effects of a relative small set of these residues, the predicted structure will be different. (b) Some other residues are less important. Despite deleting a large set of these residues, the protein still folds into a similar structure. (c) Some substitutions are radical to the protein structure and even a small number of such substitutions can drastically change the structure. (d) Some other substitutions are conservative and have small effect on the protein structure.}
    \label{fig:overview}
\vspace{-10pt}
\end{figure*}

It has been established that certain amino acids play significant roles in the protein folding process. For instance, one single disorder in the HBB gene can significantly change the structure of hemoglobin, the protein that carries oxygen in blood, causing the sickle-cell anaemia \cite{kato2018sickle}. Knowing the relationship between amino acids and protein structure helps scientists to produce synthetic proteins with precisely controlled structures \cite{tan2020chemical} or modify existing proteins with desired properties \cite{szymkowski2005creating, martinez2020unveiling, ackers1985effects}, which are essential for advanced research directions such as drug design. Additionally, in certain research tasks, scientists would like to modify the amino acids without drastically changing the protein structure, which requires the knowledge of ``safe'' residue substitutions \cite{bordo1991suggestions}, i.e., the knowledge of which amino acids are not the most crucial ones in the folding process. 

While currently there are few Explainable AI-based methods to study the mechanism of protein folding, many previous biochemical studies have been conducted for this purpose. One of the best known methods is via site-directed mutagenesis \cite{hutchison1978mutagenesis, carter1986site, sarkar1990megaprimer}. To test the role of certain residues in protein folding, biologists either delete them from the sequence (i.e., site-directed deletion) \cite{arpino2014random, gluck2002analysis, flores2007effect, dominy2003site} or replace them with other types of amino acids (i.e., site-directed substitution) \cite{flores2007effect,bordo1991suggestions, betts2003amino} and then measure their influences on the 3D structure. However, these approaches suffer from several limitations: 1) So far, modification of such residues can be limited by methods for their installation and the chemistry available for reaction, and the modification of some residues can be very challenging \cite{spicer2014selective},
2) Wet-lab methods for determining protein structures are very difficult and time-consuming \cite{ilari2008protein}, and 3) The wet lab experiments described above have many prerequisites and obstacles, and may not be completely safe for many researchers. 

Recently, AI-based dry-lab methods such as AlphaFold provide near-experimental protein structure predictions \cite{jumper2021highly}, which sheds light on the possibility to generate insightful understandings of protein folding by explaining AlphaFold's inference process. Such (Explainable) AI-driven dry-lab approach will largely overcome the aforementioned limitations and can be very helpful for human scientists. Fortunately, we observe that the process of testing the effects of residues on protein structure by site-directed mutagenesis is fundamentally similar to counterfactual reasoning, a commonly used technique for generating explanations for machine learning models \cite{tan2021counterfactual,tan2022learning,goyal2019counterfactual,tolkachev2022counterfactual,cito2022counterfactual}. Intuitively, counterfactual reasoning perturbs parts of the input data, such as interaction records of a user \cite{tan2021counterfactual}, nodes or edges of a graph \cite{tan2022learning}, pixels of an image \cite{goyal2019counterfactual}, or words of a sentence \cite{tolkachev2022counterfactual}, and then observes how the model output changes accordingly.

In this paper, we propose ExplainableFold, a counterfactual explanation framework that generates explanations for protein structure prediction models. ExplainableFold mimics existing biochemical experiments by manipulating the amino acids in a protein sequence to alter the protein structure through carefully designed optimization objectives. It provides insights about which residue(s) of a sequence is crucial (or indecisive) to the protein's structure and how certain changes on the residue(s) will change the structure, which helps to understand, e.g., what are the most impactful amino acids on the structure, and what are the most radical (or safe) substitutions when modifying a protein structure. An example of applying our framework on CASP14 target protein T1030 is shown in Figure \ref{fig:overview}, which shows that deletion or substitution of a small number of residues can result in significant changes to the protein structure, while some other deletions or substitutions may have very small effects. We evaluate the framework based on both standard explainable AI metrics and biochemical heuristics. Experiments show that the proposed method produces more faithful explanations compared to previous statistical baselines. Meanwhile, the predicted relationship between amino acids and protein structures are highly positively correlated with wet-lab biochemical experimental results.

\section{Related Work}
The essential idea of the proposed method is to integrate counterfactual reasoning and site-directed mutagenesis analysis in a unified machine learning framework. We discuss the two research directions in this section.

\subsection{Residue Effect Analysis by Site-directed Mutagenesis}
Many studies in molecular biology, such as those involving genes and proteins, rely on the use of human-induced mutation analysis \cite{stenson2017human}. Early mutagenesis methods were not site-specific, resulting in entirely random and indiscriminate mutations \cite{10.1039/9781847555380}. In 1978, \citeauthor{hutchison1978mutagenesis} \cite{hutchison1978mutagenesis} proposed the first method that modifies biological sequences at desired positions with specific intentions, known as site-directed mutagenesis. Later, more precise and effective tools have been developed \cite{motohashi2015simple, doering2018silico}. Site-directed mutagenesis is widely utilized in biomedical research for various applications. In this section, we focus on the use of site-directed mutagenesis to study the impact of amino acid mutations on protein structures \cite{studer2013residue}. 

Two common approaches to site-directed mutagenesis are amino acid deletion and substitution \cite{choi2015provean}. The deletion approach involves the deletion of certain residues from the sequence and observes the effects on the structure. For instance, \citet{gluck2002analysis} identified the amino acids that are essential to the action of the ribotoxin restrictocin by systematic deletion of its amino acids. \citet{flores2007effect} proposed a random deletion approach to measure the amino acids' effects on the longest loop of GFP. \citet{arpino2014random} conducted experiments to measure the protein's tolerance to random single amino acid deletion. The substitution approach, on the other hand, replaces one or multiple residues with other types of amino acids to test their influence. For example, \citet{clemmons2001use} substituted a small domain of the IGF-binding protein to measure whether specific domains account for specific structures and functions. \citet{zhang2018propagated} mutated a specific amino acid on the surface of a Pin1 sub-region, known as the WW domain, and observed significant structural change on the protein structure. \citet{guo2004protein} randomly replaced amino acids to test proteins' tolerance to substitution at different positions. 

When developing our framework, we draw insights from the aforementioned biochemical methods, which were proven effective in wet-lab experiments. We aim to translate the wet-lab methods of understanding protein structures into a dry-lab AI-driven approach. We note that there have been existing attempts which built models to understand the relationship between protein structures and their residues \cite{masso2008accurate, masso2006computational, sotomayor2022linking}. However, they were mostly based on statistical analysis on wet-lab experimentation data. Our method is the first AI-driven machine learning method developed for understanding protein structure predictions.

\subsection{\mbox{Counterfactual Reasoning for Explainable AI}}
Counterfactual explanation is a type of model-agnostic explainable AI method that tries to understand the underlying mechanism of a model's behavior by perturbing its input. The basic idea is to investigate the difference of the model's prediction before and after changing the input data in specific ways \cite{wachter2017counterfactual}. Since counterfactual explanation is well-suited for explaining black-box models, it has been an important explainable AI method and has been employed in various applications, such as recommender system \cite{tan2021counterfactual}, 
computer vision \cite{goyal2019counterfactual, vermeire2022explainable}, 
natural language processing \cite{yang2020generating, lampridis2020explaining,tolkachev2022counterfactual}, graph and molecular analysis \cite{tan2022learning, lin2021generative}, and software engineering \cite{cito2022counterfactual}.

In this paper, we explore counterfactual explanation to explain the amino acids' effects on protein folding. However, counterfactual explanation for protein folding has unique challenges compared with previous tasks. For example, 1) most of the aforementioned applications are classification tasks, for which the explanation goal is very clear---find a minimal change on the input that alters the predicted label. However, protein structure prediction is a generation task in a continuous space, which requires careful design of the counterfactual reasoning objective; 2) protein structure prediction models such as AlphaFold take complicated input besides the primary sequence, e.g., the MSA and templates; 3) it is easier to evaluate the explanations for the classification tasks, nevertheless, as a new AI task, protein structure prediction poses unique challenges on the evaluation of explanation. We will show how to overcome these challenges in the following parts of the paper.

\section{Problem Formulation}\label{sec:problem}
In this section, we first provide formulation of the ExplainableFold problem. Given that a protein tertiary (3D) structure is uniquely determined by its primary structure (amino acid sequences) \cite{dill2008protein, wiltgen2009structualbio}, according to the key idea of counterfactual explanation, we define the explanation as identifying the most crucial residues that cause the proteins to fold into the structures they are.

Suppose a protein consists of a chain of $l$ residues, where the $i$-th residue is encoded as a $21$-dimensional one-hot column vector $\bm{r}_i$. The ``$1$'' element in $\bm{r}_i$ indicates the type of the residue, which can be one of the $20$ common amino acids or an additional dimension for unknown residue. By concatenating all the residue vectors, a protein $P$ is denoted as 
$\bm{P}=[\bm{r}_1, \bm{r}_2, \cdots, \bm{r}_l]$, where $\bm{P} \in \{0, 1\}^{21 \times l}$ is called the protein embedding matrix. 
Many state-of-the-art protein structure prediction models predict the 3D structure not only based on the residue sequence, but also utilize supplementary evolutionary information \cite{senior2020improved, jumper2021highly} by extracting Multiple Sequence Alignment (MSA) \cite{edgar2006multiple} from protein databases. Suppose $m$ proteins are retrieved from the evolutionary database based on their similarity with protein $P$, the constructed MSAs can be encoded as another matrix $\bm{M}(\bm{P}) \in \{0, 1\} ^ {m \times 21 \times l}$. A protein structure prediction model $f_\theta$ predicts the protein 3D structure $\bm{S}$ based on the residue sequence and MSA embeddings:
\begin{equation}
    \bm{S} = f_\theta\big(\bm{P}, \bm{M}(\bm{P})\big)
\end{equation}
where $\bm{M}(\bm{P})$ can be omitted if the model only takes the residue sequence information. Though a structure prediction model may predict the positions of all atoms, in many structural biology research, only the backbone of residues are used for comparing the similarities of protein structures \cite{zhang2004scoring, zhang2005tm, xu2010significant, zemla2003lga}. Therefore, we adopt the same idea in this paper, where $\bm{S} \in \mathbb{R}^{3\times l}$ only contains the predicted $(x, y, z)^T$ coordinates of the $\alpha$-carbon atom of each amino acid residue.

The explanation is expected to be a subset of residues extracted from the protein sequence, expressed as $\mathcal{E}$. The objective of the ExplainableFold problem is to find the \textbf{minimum} set of $\mathcal{E}$ that contains the \textbf{most influential} information for the prediction of the 3D structure.

\section{The ExplainableFold Framework}
\label{sec:framework}
In biochemistry, the most common methods for studying the effects of amino acids on protein structure fall into two categories: amino acid deletion and substitution \cite{choi2015provean}. We design the ExplainableFold framework from both of the two perspectives, and we introduce them separately in the following.

\subsection{The Residue Deletion Approach}
The deletion approach simulates the biochemical studies that detect essential residues for a protein by deleting one or more residues and measuring the protein's tolerance to such deletion \cite{arpino2014random, gluck2002analysis, flores2007effect}. The key idea is to apply a residue mask that removes the effect of certain residues from the sequence and then measure the change of the protein structure. From the counterfactual machine learning perspective, this can be considered from two complementary views \cite{guidotti2019factual,tan2022learning}: 1) Identify the minimal deletion that will alter the predicted structure and the deleted residues will be the \textit{necessary} explanation; 2) Identify the maximal deletion that still keeps the predicted structure and the undeleted residues will be the \textit{sufficient} explanation. We design the counterfactual explanation algorithm from these two views accordingly.

\subsubsection{\textbf{Necessary Explanation (Most Intolerant Deletion)}}
From the necessary perspective, we aim to find the \textbf{minimal} set of residues in the original sequence which, if deleted, \textbf{will change} the AI model's (such as  AlphaFold's) predicted structure. The \textbf{deleted} residues thus contain the most necessary information for the model's original prediction.

We can express the perturbation on the original sequence as a multi-hot vector $\Delta = \{0, 1\}^{1\times l}$, where $\delta_i=1$ means that the $i$-th residue will be deleted and $\delta_i=0$ means it will be kept. Then the counterfactual protein embedding matrix $\bm{P^}\Delta$ can be expressed as:
\begin{equation}
\begin{aligned}
    & \bm{P}^\Delta = \bm{P} \odot (\mathbf{1} - \Delta) + \bm{U} \odot \Delta
\end{aligned}
\end{equation}
where $\odot$ is the element-wise product, $\mathbf{1}$ is an all-1 vector with length $l$, and $\bm{U} \in \{0, 1\}^{21 \times l}$ denotes an ``unknown'' matrix of the same shape with $\bm{P}$, but with all elements being $0$ except for the last row being $1$ (i.e., unknown type amino acid). Thus, for $\delta_i = 0$, the $i$-th residue in the original sequence will be preserved, while for $\delta_i = 1$, the $i$-th residue will be treated as an unknown amino acid without any specific chemical property. 

Motivated by the Occam's Razor Principle \cite{blumer1987occam}, we aim to find \textit{simple} and \textit{effective} explanations. The simpleness can be characterized by the number of residues that need to be deleted, which should be as few as possible, while effectiveness means that the predicted protein structure should be different before and after applying the deletions. We can use zero-norm $\|\Delta\|_0$ to represent the number of deletions (for simpleness), while using the TM-score between the original and the new protein structures $\text{TM}(\bm{S}, \bm{S^*})$ to represent the degree of change on the structure (for effectiveness). TM-score is a standard measurement for comparing aligned protein structures, where TM-score $> 0.5$ suggests the same folding and TM-score $\le 0.5$ suggests different foldings \cite{zhang2004scoring, xu2010significant}. The counterfactual explanation algorithm then learns the optimal explanation by solving the following constrained optimization problem:
\begin{equation}
\label{eq:deletion_necessary_discrete}
    \begin{aligned}
        &\minimize \|\Delta\|_0 \\
        &~\text{s.t.} ~~\text{TM}(\bm{S}, \bm{S^*}) \leq 0.5, ~~\Delta = \{0, 1\}^{1\times l} \\
        &~\text{where} ~~\bm{S^*}=f_\theta(\bm{P}^\Delta, \bm{M}(\bm{P}^\Delta))
    \end{aligned}
\end{equation}
where the objective $\|\Delta\|_0$ aims to find the minimal deletion, while the constraint guarantees the effectiveness of the deletion, i.e., the deletion will change the predicted protein structure to be different from before.

Due to the exponential combinations of sub-sequences for a given sequence, it is impractical to search for an optimal solution on the discrete space. To solve the problem, we use a continuous relaxation approach to solve the optimization problem by relaxing the multi-hot vector $\Delta$ to a real-valued vector. We also relax the hard constraint in Eq.\eqref{eq:deletion_necessary_discrete} and combine them into a single trainable loss function: 
\begin{equation}
\label{eq:deletion_necessary_loss}
\begin{aligned}
    &\mathcal{L}_1= \text{LeakyReLU}\big (\text{TM}(\bm{S}, \bm{S^*}) - 0.5 + \alpha \big ) + \lambda_1 \|\sigma(\Delta)\|_1\\
    &\text{s.t.} ~\Delta \in \mathbb{R}^{1\times l},\text{where}~\bm{S^*}=f_\theta(\bm{P}^{\sigma(\Delta)}, \bm{M}(\bm{P}^{\sigma(\Delta)}))
\end{aligned}
\end{equation}
where the LeakyReLU function is $\text{LeakyReLU}(x) = \max(0,x) + \text{negative\_slope} \cdot \min(0,x)$ \cite{maas2013rectifier} and we set $\text{negative\_slope} = 0.1$, the sigmoid function $\sigma(\cdot)$ is applied so that $\sigma(\Delta) \in (0,1)^{1\times l}$ approximates the probability distribution between the original residues and unknown residues, and $\alpha$ is the margin value whose default value is $0.1$. This relaxation approach has been justified in several previous studies which also learn explanation on discrete inputs \cite{goyal2019counterfactual, tan2022learning}. The $1$-norm regularizer assures the learned perturbation $\sigma(\Delta)$ to be sparse \cite{candes2005decoding}, i.e., the learned explanation only contains a small set of residues. $\lambda_1$ is a hyper-parameter that controls the trade-off between the complexity and strength of the generated explanation.

We use the LeakyReLU loss function as a variation of the original hinge loss function $\text{hinge}(x)=\max(0,x)$ to make the loss optimization process more stable and smooth, which will be explained in the implementation details subsection \ref{sec:implementation_details}. Eq.\eqref{eq:deletion_necessary_loss} can be easily optimized with gradient descent. After optimization, we convert $\sigma(\Delta)$ to a binary vector with the threshold of $0.5$. 

\subsubsection{\textbf{Sufficient Explanation (Most Tolerant Deletion)}}
\label{section:sufficient_tolerant}
Symmetrically, from the sufficiency perspective, we aim to find the \textbf{maximal} set of residues in the orignal sequence which, if deleted, \textbf{will not change} the AI model's predicted structure. The \textbf{undeleted} residues thus contain the most sufficient information for the model's original prediction.

This can be formulated as a similar but reversed optimization process as Eq.\eqref{eq:deletion_necessary_discrete}, which looks for the maximal perturbation $\Delta$ while keeping the same folding (TM-score $>0.5$). Therefore, the optimization problem is formulated as:
\begin{equation}
\label{eq:deletion_sufficient_discrete}
    \begin{aligned}
        &\maximize \|\Delta\|_0 \\
        &~\text{s.t.} ~~\text{TM}(\bm{S}, \bm{S^*}) > 0.5, ~~\Delta = \{0, 1\}^{1\times l} \\
        &~\text{where} ~~\bm{S^*}=f_\theta\big(\bm{P}^\Delta, \bm{M}(\bm{P}^\Delta)\big)
    \end{aligned}
\end{equation}
Similarly, we relax Eq.\eqref{eq:deletion_sufficient_discrete} to a differentiable loss function:
\begin{equation}
\label{eq:deletion_sufficient_loss}
\begin{aligned}
    &\mathcal{L}_2= \text{LeakyReLU}\big(0.5 - \text{TM}(\bm{S}, \bm{S^*}) + \alpha \big) - \lambda_2 \|\sigma(\Delta)\|_1\\
    &\text{s.t.} ~\Delta \in \mathbb{R}^{1\times l},\text{where} ~\bm{S^*}=f_\theta(\bm{P}^{\sigma(\Delta)}, \bm{M}(\bm{P}^{\sigma(\Delta)}))
\end{aligned}
\end{equation}
Contrary to the necessary explanation, the sufficient explanation consists the undeleted residues. Hence, after optimization, we filter the residues according to $\big(1-\sigma(\Delta)\big)>0.5$ and include them into the sufficient explanation.

\subsection{The Residue Substitution Approach}
Another popular approach in biochemistry, site-directed substitution, studies the influence of the amino acids on protein folding by replacing certain residues with other known-type residues \cite{flores2007effect,bordo1991suggestions, betts2003amino}. Different replacements may have different effects on protein structures, and they can be classified into two types: conservative substitution and radical substitution \cite{zhang2000rates, dagan2002ratios}. A conservative substitution is considered as a ``safe'' substitution for which the amino acid replacement usually have no or minor effects on the protein structure. A radical substitution is considered ``unsafe,'' which usually causes significant structural changes. Based on the above concepts, we design the substitution approach from these two different perspectives.

\subsubsection{\textbf{Radical Substitution Explanation}} From the radical substitution perspective, we aim to find the \textbf{mimimal} set of residue replacements which will lead to a \textbf{different} folding, and then the learned substitutions are the most \textbf{radical} substitutions for the protein. 

For a target protein with binary embedding matrix $\bm{P}$, we learn a counterfactual binary protein embedding $\bm{P^\prime}$, which has the same shape as the original embedding matrix. The number of substitutions is represented by $\|\bm{P} - \bm{P^\prime}\|_0$, which is the $0$-norm of the difference between the two matrices. To find the minimal residue substitution that changes the original folding, the optimization problem is defined as:
\begin{equation}
\label{eq:substitution_radical_discrete}
    \begin{aligned}
        & \minimize ||\bm{P}-\bm{P^\prime}||_0\\
        &~\text{s.t.} ~\text{TM}(\bm{S}, \bm{S^\prime}) \leq 0.5, ~~\bm{P^\prime} \in \{0, 1\}^{21\times l}\\
        &~\text{where} ~~\bm{S^\prime}=f_\theta\big(\bm{P^\prime}, \bm{M}(\bm{P^\prime})\big)
    \end{aligned}
\end{equation}
Due to the exponential search space of the substitutions, we use the similar continuous relaxation method as in the deletion approach. First, we relax the binary counterfactual embedding matrix $\bm{P^\prime}$ to continuous space. We also relax the hard constraint in Eq.\eqref{eq:substitution_radical_discrete} and define the differentiable loss function as:
\begin{equation}
\begin{aligned}
\label{eq:substitution_radical_discrete_loss}
    &\mathcal{L}_3 = \text{LeakyReLU} \big (\text{TM}(\bm{S}, \bm{S^\prime}) - 0.5 + \alpha \big ) + \lambda_3 \|\bm{P}-\sigma(\bm{P^\prime})\|_1 \\
    & \text{s.t.} ~\bm{P^\prime}\in \mathbb{R}^{21\times l},\text{where} ~\bm{S^\prime}=f_\theta\big(\bm{P^\prime}, \bm{M}(\bm{P^\prime})\big)
\end{aligned}
\end{equation}
After optimization, we convert the learned continuous matrix $\sigma(\bm{P^\prime})$ into binary by setting the maximum value of each column as 1 and others as 0. Then, the changed residues between $\bm{P}$ and $\bm{P^\prime}$ are the radical substitution explanations.

\subsubsection{\textbf{Conservative Substitution Explanation}}
From the conservative substitution perspective, we aim to find the \textbf{maximal} set of residue replacements which however lead to the \textbf{same} folding, and then the learned substitutions are the most \textbf{conservative} substitutions for the protein.

On the contrary to Eq.\eqref{eq:substitution_radical_discrete}, we formulate an inverse optimization problem as:
\begin{equation}
\label{eq:substitution_conservative_discrete}
    \begin{aligned}
        & \maximize ||\bm{P}-\bm{P^\prime}||_0\\
        &~\text{s.t.} ~\text{TM}(\bm{S}, \bm{S^\prime}) > 0.5, ~~ \bm{P^\prime} \in \{0,1\}^{21 \times l}\\
        &~\text{where} ~~\bm{S^\prime}=f_\theta\big(\bm{P^\prime}, \bm{M}(\bm{P^\prime})\big)
    \end{aligned}
\end{equation}
With the same relaxation process, the loss function is:
\begin{equation}
\begin{aligned}
\label{eq:substitution_conservative_discrete_loss}
    &\mathcal{L}_4 = \text{LeakyReLU} \big(0.5-\text{TM}(\bm{S}, \bm{S}^\prime)+\alpha \big)- \lambda_4\|\bm{P}-\sigma(\bm{P^\prime})\|_1 \\
    & \text{s.t.} ~\bm{P^\prime}\in \mathbb{R}^{21\times l},\text{where} ~\bm{S^\prime}=f_\theta\big(\bm{P^\prime}, \bm{M}(\bm{P^\prime})\big)
\end{aligned}
\end{equation}
After learning $\sigma(\bm{P^\prime})$ and getting the binary matrix, again, the changed residues between $\bm{P}$ and $\bm{P^\prime}$ are the conservative substitution explanations.

\subsection{Phased MSA Re-alignment}
It is impractical to re-compute MSAs in each training step. Therefore, we propose a phased MSA re-alignment strategy. When learning the explanations, we fix the generated MSAs and only learn the changes on the sequence embedding for $t$ training steps ($t=100$ by default), which is one phase. Then, we re-align the MSAs and start another training phase. 

\section{Experiments}
\label{sec:experiments}
We first introduce the datasets and implementation details. Then, we introduce the evaluation results of the deletion approach and substitution approach, respectively. 

\subsection{Datasets}
We test the ExplainableFold framework on the 14th Critical Assessment of protein Structure Prediction (CASP-14) dataset\footnote{\url{https://predictioncenter.org/casp14/}} \cite{moult14critical}. CASP consecutively establishes protein data with detailed structural information as a standard evaluation benchmark for protein structure prediction. Following \citet{jumper2021highly}, we remove all sequences for which fewer than $80$ amino acids had the alpha carbon resolved and remove duplicated sequences. 

\begin{table*}[t]
\caption{PN Evaluation. Deletion$^*$ is the necessity optimization.}
\vspace{-5pt}
\centering
% \begin{adjustbox}{width=0.7\linewidth}
\begin{tabular}{lrrrr}
\toprule
& Ave Explanation & Ave Complexity & Ave TM-score & PN\\ 
& Size ($|\mathcal{E}|$) $\downarrow$ & ($|\mathcal{E}|/l$) $\downarrow$ & $\text{TM}(\bm{S}, \bm{S^*})\downarrow$ & score$\uparrow$\\
\cmidrule(lr){1-5}
Random    & 85.22           & 0.33         & 0.83             & 0.07   \\
Evolutionary \cite{masso2006computational}    & 88.42           & 0.33         &  0.77            & 0.16   \\
Deletion (necessity)$^*$ & \textbf{83.33}           & \textbf{0.31}         & \textbf{0.59}             & \textbf{0.40}   \\
\bottomrule
\end{tabular}
% \end{adjustbox}
\label{tab:pn_evaluation}
% \vspace{-10pt}
\end{table*}

\begin{table*}[t]
\caption{PS Evaluation. Deletion$^*$ is the sufficiency optimization.}
\vspace{-5pt}
\centering
% \begin{adjustbox}{width=0.7\linewidth}
\begin{tabular}{lrrrr}
\toprule
& Ave Explanation & Ave Complexity & Ave TM-score & PS\\ 
& Size ($|\mathcal{E}|$) $\downarrow$ & ($|\mathcal{E}|/l$) $\downarrow$ & $\text{TM}(\bm{S}, \bm{S^*})\uparrow$ & score$\uparrow$\\
\cmidrule(lr){1-5}
Random    & 129.78           & 0.50          & 0.44             & 0.36   \\
Evolutionary \cite{masso2006computational}  & 134.89 & 0.51 & 0.49 & 0.42 \\
Deletion (sufficiency)$^*$ &     \textbf{106.25}       & \textbf{0.49}         &  \textbf{0.65}            & \textbf{0.76}   \\
\bottomrule
\end{tabular}
% \end{adjustbox}
\label{tab:ps_evaluation}
% \vspace{-10pt}
\end{table*}

\subsection{Implementation Details}
\label{sec:implementation_details}
Though the ExplainableFold framework can be applied on any model that predicts protein 3D structures, we choose Alphafold2 \cite{jumper2021highly}, the state-of-the-art model, as the base model in the experiments. More specifically, we use the OpenFold \cite{Ahdritz2022.11.20.517210} implementation, and load the official pre-trained AlphaFold parameters\footnote{\url{https://github.com/deepmind/alphafold}}. 

When learning the explanations, the pre-trained parameters of AlphaFold are fixed, and only the perturbation vectors on the input ($\Delta$ for the deletion approach and $\bm{P}^\prime$ for the substitution approach) will be optimized. However, it still requires computing the gradient through the entire Alphafold network, as a result, the learning process requires extensive memory consumption. To solve the problem, we follow exactly the same training procedure as introduced in the original AlphaFold paper \cite{jumper2021highly}. More specifically, we use the gradient checkpointing technique to reduce the memory usage \cite{chen2016training}. Meanwhile, if a protein has more than $384$ residues, we cut it to different chunks for each consecutive $384$ residues and generate explanations for each chunk \cite{jumper2021highly}. Except for memory effciency, this also makes it meaningful to compute and compare the explanation size since the maximum sequence length is bounded.

We employ the same training strategy for both deletion and substitution explanation methods: for each training phase between MSA re-alignments, we optimize the perturbation vector for $100$ steps with Adam optimizer \cite{kingma2014adam} and learning rate $0.01$. After each training loop, we re-align the MSAs with the AlphaFold HHblits / JackHMMER pipeline. We repeat the training and alignment process for $3$ phases when generating explanations for each protein. All experiments are conducted on NVIDIA A5000 GPUs. The entire training process, including all $3$ phases, for one protein takes approximately $5$ hours. We set $\alpha=0.2$ and $\lambda=0.00001$, $\lambda=0.002$, $\lambda=0.01$, $\lambda=0.0001$ in Equations \eqref{eq:deletion_necessary_loss}\eqref{eq:deletion_sufficient_loss}\eqref{eq:substitution_radical_discrete_loss}\eqref{eq:substitution_conservative_discrete_loss}, respectively. To realize an incremental deletion/substitution process, we initialize the counterfactual protein embedding matrix as a near duplication of the original protein embedding matrix, i.e., we initialize $\Delta$ with near $0$'s and initialize $\bm{\sigma(P^\prime)}$ approximately equal to the original $\bm{P}$. 

However, the above initialization may lead to unstable optimization if we use hinge loss as the loss function. Take Eq.\eqref{eq:deletion_sufficient_loss} as an example, the initial TM score is close to 1 under this initialization, making the value of $0.5 - \text{TM}(\bm{S}, \bm{S^*}) + \alpha$ negative. As a result, if we use hinge loss in Eq.\eqref{eq:deletion_sufficient_loss}, i.e., $\max\big(0, 0.5 - \text{TM}(\bm{S}, \bm{S^*}) + \alpha \big) - \lambda_2 \|\sigma(\Delta)\|_1$, then the hinge part of the loss will not take any effect during early phase of the optimization, making the optimization process unstable. Therefore, we use the LeakyReLU function as a varianiton of the original hinge loss function to ensure a stable learning process.

\subsection{Evaluation of the Deletion Approach}
Counterfactual explanations can be evaluated by their complexity, sufficiency and necessity \cite{glymour2016causal,tan2022learning}. First, according to the Occam's Razor Principle \cite{blumer1987occam}, we hope an explanation can be as simple as possible so that it is cognitively easy to understand for humans. This can be evaluated by the complexity of the explanation, i.e., the percentage of residues that are included in the explanation:

\begin{equation}
    \text{Complexity} = |\mathcal{E}| / l
\end{equation}
where $l$ is the length of the protein.

Sufficiency and necessity measure how crucial the generated explanations are for the protein structure. We follow the definition in causal inference theory \cite{glymour2016causal} and existing explainable AI research \cite{tan2022learning} and measure the explanations with two causal metrics: Probability of Necessity (PN) and Probability of Sufficiency (PS). 

PN measures the necessity of the explanation. A set of explanation residues is considered a necessary explanation if, by removing their effects from the protein sequence, the predicted structure of the protein will have a different folding (TM-score $<0.5$). Suppose there are $N$ proteins in the testing data, then PN is calculated as:
\begin{equation}
\begin{aligned}
% \small
\hspace{-4pt}
    \text{PN} = \frac{\sum_{k=1}^N \text{PN}_k}{N},~
    \text{PN}_k=\begin{cases}
      1,\text{if} ~\text{TM}(\bm{S}_k, \bm{S^*}_k) \leq 0.5\  \\
      0,\text{else}
    \end{cases}
\end{aligned}
\end{equation}
Intuitively, PN measures the percentage of proteins whose explanation residues, if removed, will change the protein structure, and thus their explanation residues are necessary.

PS measures the sufficiency of the explanation. A set of explanation residues is considered a sufficient explanation if, by removing all of the other residues and only keeping the explanation residues, the protein still has the same folding. Similarty, PS is calculated as:
\begin{equation}
\begin{aligned}
% \small
\hspace{-4pt}
    \text{PS} = \frac{\sum_{k=1}^N \text{PS}_k}{N},
    ~\text{PS}_k=\begin{cases}
      1,\text{if} ~\text{TM}(\bm{S}_k, \bm{S^*}_k) > 0.5\  \\
      0,\text{else}
    \end{cases}
\end{aligned}
\end{equation}
Intuitively, PS measures the percentage of proteins whose explanation residues alone can keep the protein structure unchanged, and thus their explanation resides are sufficient.

\subsubsection{\textbf{Baselines}}
\begin{figure*}[t]
% \vspace{-10pt}
\centering
% \hspace{-10pt}
\mbox{
    \centering
    \subfigure[Top 30\% conserved residues]{
        \includegraphics[width=0.35\textwidth]{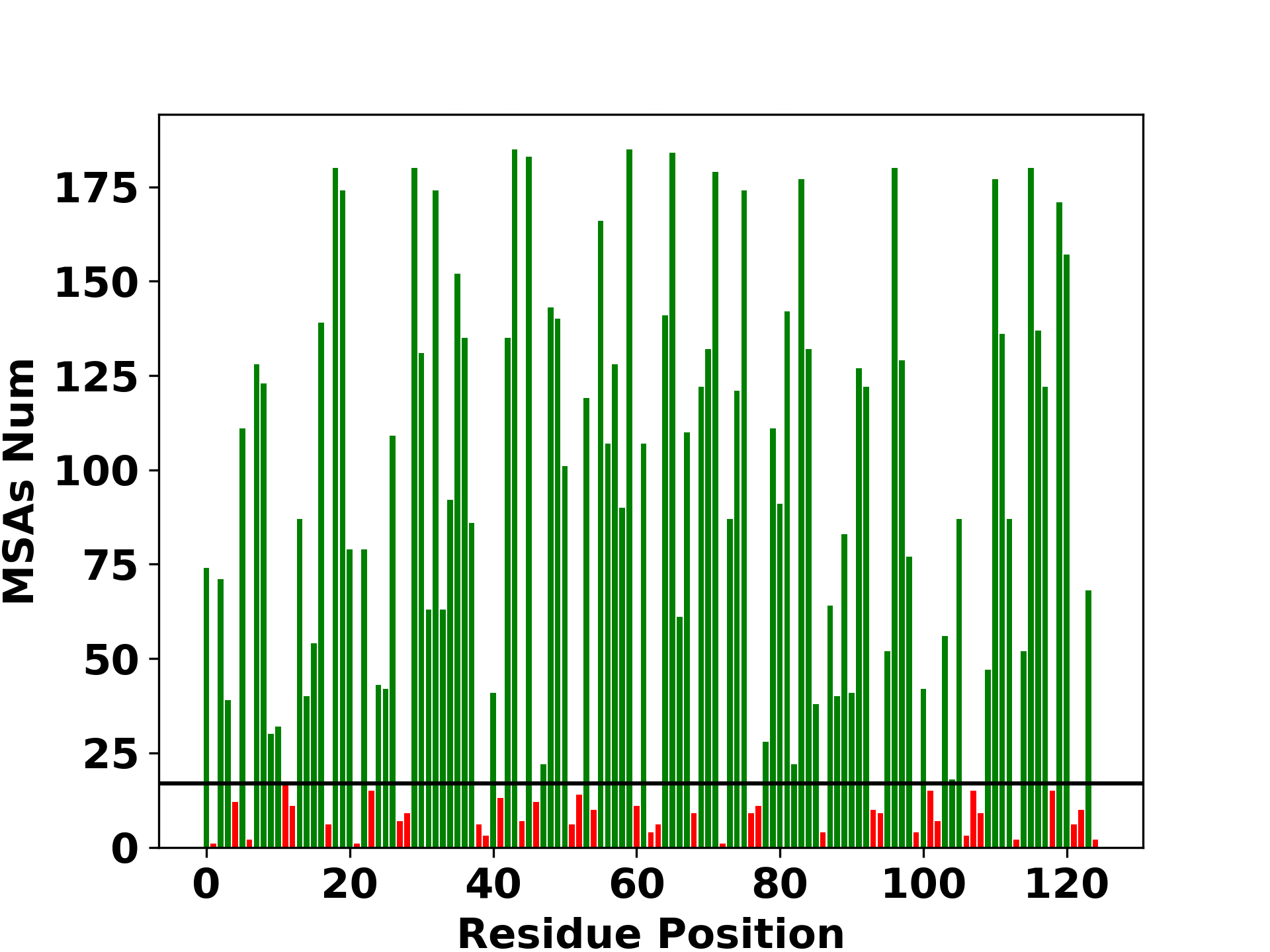}}
    % \hspace{-10pt}
    \subfigure[Top 40\% conserved residues]{
        \includegraphics[width=0.35\textwidth]{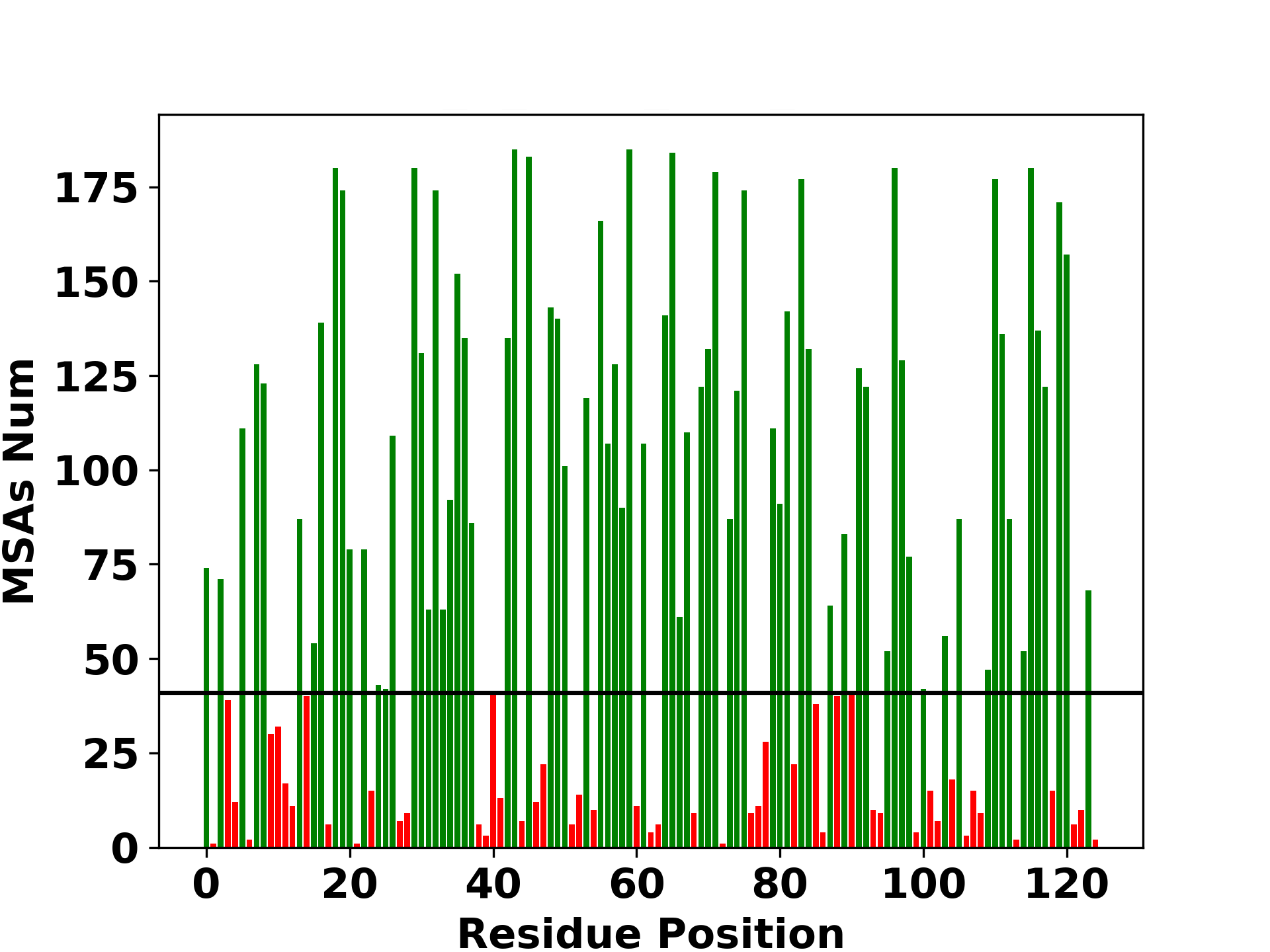}}
}
\vspace{-10pt}
\caption{Evolutionary conserved residues are considered more important for the protein structure (the residues in red).}
\label{fig:evol_baseline}
\vspace{-15pt}
\end{figure*}

We compare the model performance with a common computational biology baseline \cite{masso2006computational}, which analyzes a protein's tolerance to the change on each residue by extracting the data from evolutionary database. More specifically, proteins are not tolerant to the mutations at evolutionarily conserved positions. However, they are capable of withstanding certain mutations at other positions. When implementing the baseline, we refer to a protein's MSAs and select the evolutionarily conserved residues as the explanation. This is illustrated in Figure \ref{fig:evol_baseline} using protein CASP14 target T1029 as an example, where for each residue position, we count the number of MSAs that conserve the residue at this position, and show the top 30\% and 40\% conserved residues. We also randomly select residues as explanation and compute PN and PS scores as another baseline to measure the general difficulty of the evaluation task, and more details are provided in the following subsection.

\begin{figure*}[t]
% \vspace{-10pt}
\centering
\mbox{
    \centering
    \subfigure[Necessary Optimization]{\includegraphics[width=0.35\textwidth]{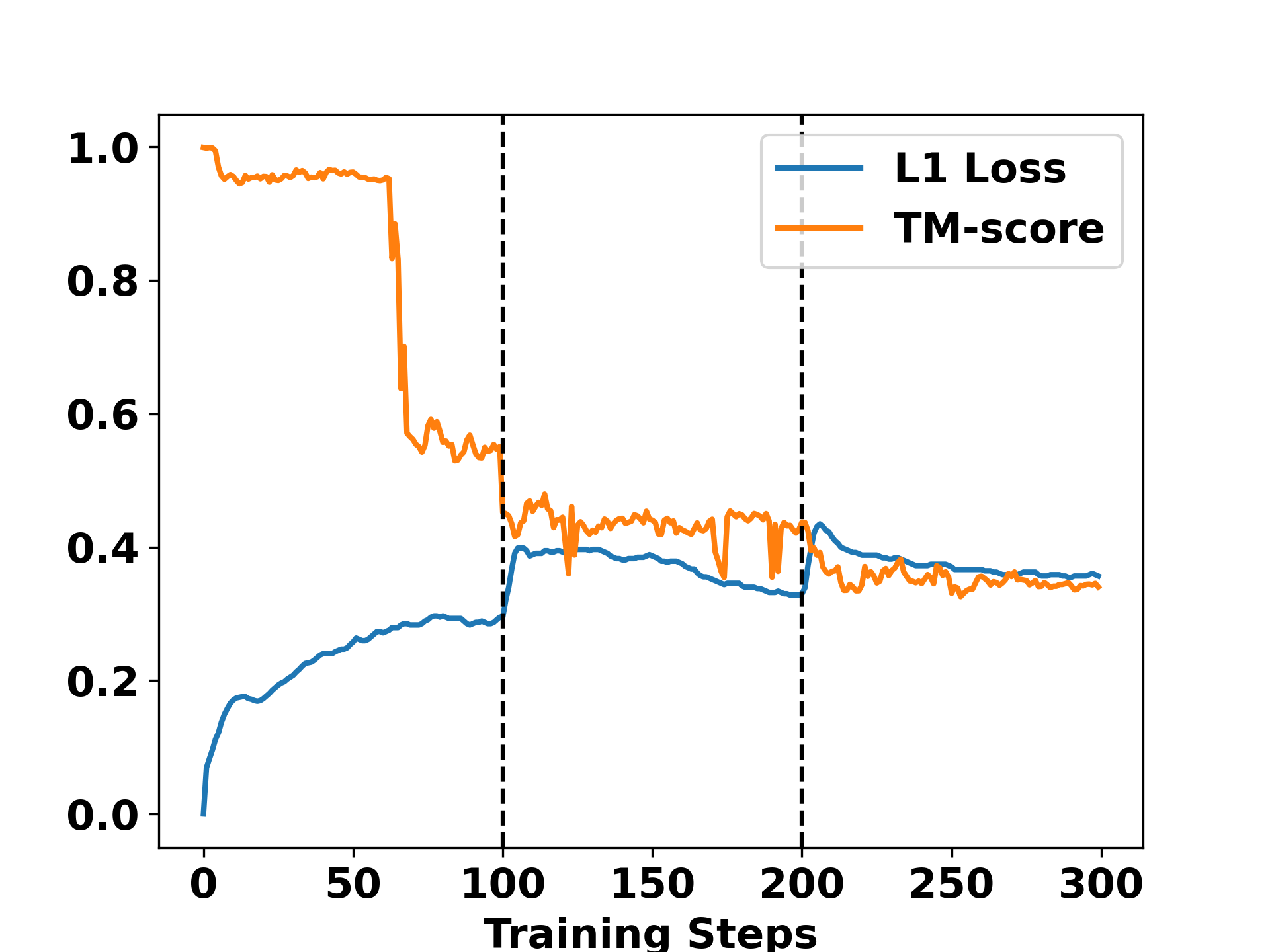}}
    % \hspace{-15pt}
    \subfigure[Sufficient Optimization]{
        \includegraphics[width=0.35\textwidth]{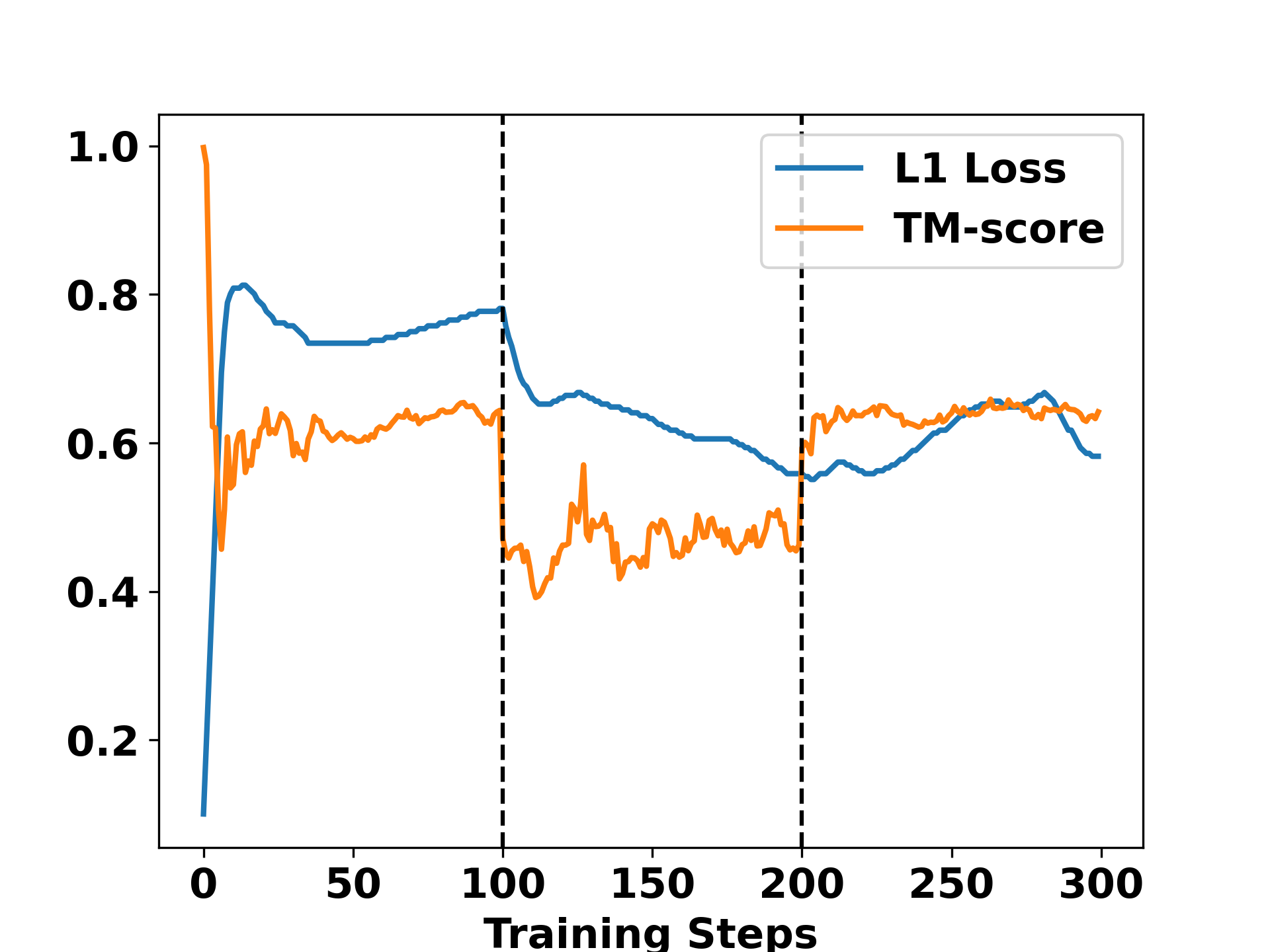}}
}
\vspace{-10pt}
\caption{Learning Curves of the Deletion Approach, including three phases between MSA re-alignment, 100 traing steps each.}
\label{fig:learning_curve}
\vspace{-5pt}
\end{figure*}

\subsubsection{\textbf{Results}}
The results of PN and PS evaluation are reported in Table \ref{tab:pn_evaluation} and Table \ref{tab:ps_evaluation}, respectively. The explanation complexities of our method ($31\%$ for necessary explanation and $49\%$ for sufficient explanation) are automatically decided by our optimization process. However, the baselines do not have the ability to decide the optimal explanation complexity. For fair comparison, we set the complexities of the baselines to be similar our method ($33\%$ for necessary explanation and $50\%$ for sufficient explanation). Therefore, the baselines will have a small advantage over our method because they are allowed to use more residues to achieve the necessity or sufficiency goals.

For PN evaluation, the results of the random baseline shows that protein structures tend to be robust to residue deletions. For example, when randomly removing the effects of 33\% residues, only $7\%$ of the proteins fold into different structures, which indicates that finding necessary explanations is a challenging problem. The evolutionary baseline is able to select more necessary residues with a PN score of $0.16$, which is $128.6\%$ better than random selection. Compared to them, our method shows much better performance: with a smaller number of residues, the generated explanations are able to cause $41\%$ of the proteins fold into different structures, outperforming the evolutionary baseline by $150\%$.

For PS evaluation, the evolutionary baseline is not noticeably better than randomly selecting residues. The reason may be that despite the proteins' less tolerance to the evolutionary conserved residues, there is no guarantee that the evolutionary conserved residues alone contain sufficient information to preserve the protein structure. In comparison, our method does generate more sufficient explanations, outperforming the evolutionary baseline by $81.0\%$ according to the PS score with less complex explanations. Meanwhile, our TM score is $>50\%$, indicating that the protein structure is indeed preserved under our sufficient explanation.

Additionally, we show the learning curve of the optimization for CASP14 target protein T1030 in Figure \ref{fig:learning_curve}. For necessary optimization, the algorithm gradually deletes the protein residues until reaching a TM-score near $0.3$ (i.e., $0.5 - \alpha$, see Eq.\eqref{eq:deletion_necessary_loss}). Then, the explanation complexity slightly drops back while keeping the TM-score at the same level. During sufficient optimization, the L1-loss drastically increases initially, which suggests that the algorithm is trying to delete as many residues as possible while keeping the original folding structure unchanged. However, after re-computing MSAs, the TM-score becomes too low. Thus, the algorithm increases the number of preserved residues to keep TM-score near $0.7$ (i.e., $0.5 + \alpha$, see Eq.\eqref{eq:deletion_sufficient_loss}). Note that the TM-scores change sharply when re-computing MSAs at the end of each training loop. More frequent MSA realignments result in a smoother optimization process.

\begin{figure*}[t]
% \vspace{-10pt}
% \hspace{-15pt}
\centering
\mbox{
    \centering
    \subfigure[Correlation with Epstein's distance]{
        \includegraphics[width=0.4\textwidth]{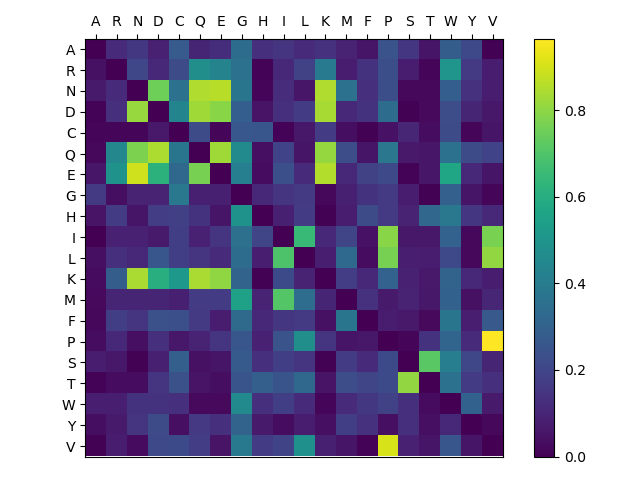}}
    % \hspace{-10pt}
    \subfigure[Correlation with Miyata's distance]{
        \includegraphics[width=0.4\textwidth]{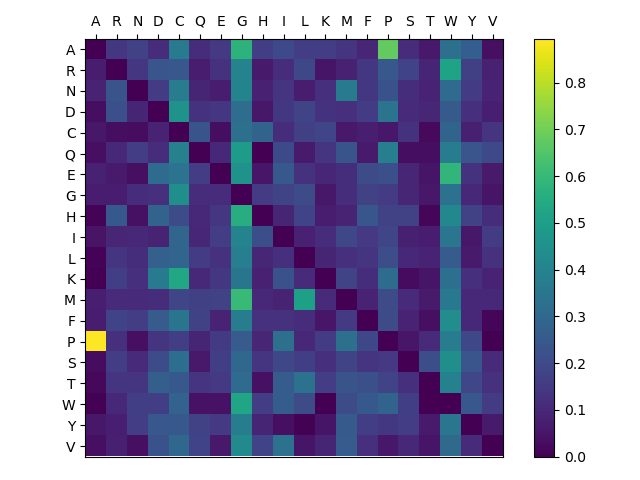}}
}
\vspace{-5pt}
\caption{The correlation between the exchangeability provided by our conservative optimization method and (a) Epstein's distance as well as (b) Miyata's distance.}
\label{fig:heatmap}
% \vspace{-10pt}
\end{figure*}

\subsection{Evaluation of the Substitution Approach}
The substitution approach identifies the most radical or conservative amino acid substitutions, which are of particular interest in biochemical research \cite{zhang2000rates}. Previously, it was impractical to conduct wet-lab experiments to investigate the relative ``safety'' of replacing specific residues with alternative amino acids due to their prohibitive cost \cite{bordo1991suggestions}. Alternatively, scientists infer the \textit{exchangeability} of two types of amino acids either through the use of heuristics based on their physical or chemical properties or through the analysis of evolutionary data, such as:
% \vspace{-1ex}
\begin{itemize}
% [leftmargin=*]
% \setlength{\parskip}{0pt}
% \setlength{\parsep}{0pt}
\item Epstein's distance \cite{epstein1967non}: Predict the impact of switching two amino acids based on their size and polarity.
\item Miyata's distance \cite{miyata1979two}: Predict the impact based on their volume and polarity.
\item Evolutionary indicator \cite{bordo1991suggestions}: Detect ``safe'' substitutions based on evolutionary data.
\end{itemize}

Note that these indicators are rather suggestions than ground-truth. They provide general trends that are better than random selection but cannot be expected to be precise in every scenario \cite{bordo1991suggestions}. These methods are not perfectly consistent with each other, but are linearly related.

Therefore, we utilize the amino acid substitution data generated by our method to caculate the pair-wise exchangeability between the amino acids, and test the correlation between our exchangeability with the above three existing exchangeability indicators. The details of the pair-wise substitution statistics and the calculation of pair-wise exchangeability are provided in the Appendix.

In Table \ref{tab:correlation}, we report the correlation of our generated pair-wise exchangeability with the three aforementioned indicators by a non-parametric method: Pearson's correlation $r$. Besides, the correlation among the three biochemical methods themselves range from $0.438$ to $0.578$. Additionally, the correlation is visualized in Figure \ref{fig:heatmap}, where darker color indicates higher correlation. For Pearson's correlation, a value greater than $0$ indicates a positive association, where $r>0.1$, $r>0.3$, $r>0.5$ represents small, medium, and large correlations, accordingly \cite{cohen2009pearson}. Both Table \ref{tab:correlation} and Figure \ref{fig:heatmap} show that our method has clear positive correlations with all of the three biochemical methods, indicating that ExplainableFold can provide informative exchangeability signals \cite{yampolsky2005exchangeability}. Besides, the results generated by ExplainableFold may further improve when larger protein datasets are available or applied on even better base models in the future.

\begin{table}[t]
\caption{Correlation between our method and each of the biochemical indicators. Metrics with ``*'' are originally distance metrics, for which we take the inverse to reprenst the exchangeability. The results are significant at $p<0.001$ under two-tailed test.}
\vspace{-5pt}
\centering
% \begin{adjustbox}{width=0.8\linewidth}
\begin{tabular}{lccc}
\toprule
& Epstein$^*$ & Miyata$^*$ & Evolutionary \\ 
% & $r$ $\uparrow$ & $r$ $\uparrow$ & $r$ $\uparrow$ \\
\cmidrule(lr){1-4}
Radical   & 0.388           & 0.602         & 0.382 \\
Conservative    & 0.494           & 0.796         & 0.405 \\
\bottomrule
\end{tabular}
% \end{adjustbox}
\label{tab:correlation}
\vspace{-10pt}
\end{table}

\section{Conclusions and Future Work}
In this paper, we propose ExplainableFold---an Explainable AI framework that helps to understand the deep learning based protein structure prediction models such as AlphaFold. Technically, we develop a counterfactual explanation framework and implement the framework based on two approaches: the residue deletion approach and the residue substitution approach. Intuitively, ExplainableFold aims to find simple explanations that are effective enough to keep or change the protein's folding structure. Experiments are conducted on CASP-14 protein datasets and results show that our approach outperforms the results from traditional biochemical methods. We believe Explainable AI is fundamentally important for AI-driven scientific research because science not only pursues the answers for the ``what'' questions but also (or even more) for the ``why'' questions. In the future, we will further improve our framework by considering more protein modification methods beyond deletion and substitution. We will also generalize our framework to other scientific problems due to the flexibility of our framework.

\section*{Acknowledgement}
We thank the reviewers for their constructive suggestions. The work was supported in part by NSF IIS 1910154, 2007907, 2046457, 2127918 and NIH CA277812-02.

%%%%%%%%%%%%%%%%%%%%%%%%%%%%%%%%%%%%%%%%%%%%%%%%%%%%%%%%%%%%%%%%%%%%%%%%%%%%%%%
%%%%%%%%%%%%%%%%%%%%%%%%%%%%%%%%%%%%%%%%%%%%%%%%%%%%%%%%%%%%%%%%%%%%%%%%%%%%%%%
% APPENDIX
%%%%%%%%%%%%%%%%%%%%%%%%%%%%%%%%%%%%%%%%%%%%%%%%%%%%%%%%%%%%%%%%%%%%%%%%%%%%%%%
%%%%%%%%%%%%%%%%%%%%%%%%%%%%%%%%%%%%%%%%%%%%%%%%%%%%%%%%%%%%%%%%%%%%%%%%%%%%%%%
% \newpage

\appendix
% \onecolumn
\section*{Appendix}
% \section{You \emph{can} have an appendix here.}
\label{sec:appendix}

\section{Statistical Analysis of Amino Acid Substitutions}
\label{sec:substitution_tables}

Table \ref{tab:total_number} shows the total count of each amino acid in the testing proteins. In Table \ref{tab:conservative_statis}, we show how many times a specific type of substitution happens in the generated explanations learned by the conservative substitution method. For instance, the substitution of $A\rightarrow R$ happens $19$ times. The exchangeability of $X \rightarrow Y$ can be easily calculated by $|X\rightarrow Y|/|X|$ \cite{bordo1991suggestions, masso2006computational}. The same statistics for radical substitution is provided in Table \ref{tab:radical_statis}. For radical substitution, the higher the number in Table \ref{tab:radical_statis}, the lower the exchangeability, and thus the exchangeability of $X\rightarrow Y$ is calculated as the reciprocal $|X|/|X\rightarrow Y|$ \cite{bordo1991suggestions, masso2006computational}.

\begin{table}[h]
\caption{Total number of each amino acid in testing data}
\vspace{-5pt}
\centering
% \begin{adjustbox}{width=.9\linewidth}
\begin{tabular}{lllllllllll}
\hline
   & A   & R   & N   & D   & C   & Q   & E   & G   & H   & I \\ \hline
\# & 782 & 581 & 827 & 776 & 175 & 520 & 848 & 853 & 309 & 904 \\ \hline
   & L   & K   & M   & F   & P   & S   & T   & W   & Y   & V   \\ \hline
\# & 1122 & 911 & 273 & 600 & 506 & 916 & 746 & 149 & 595 & 780 \\ \hline
\end{tabular}
% \end{adjustbox}
\label{tab:total_number}
\vspace{-5pt}
\end{table}

\begin{table*}[h]
\centering
\caption{Structural Conservative Statistics}
\vspace{-5pt}
\centering
% \begin{adjustbox}{width=.7\linewidth}
\begin{tabular}{l|llllllllllllllllllll}
  & A & R & N & D & C & Q & E & G & H & I & L & K & M & F & P & S & T & W & Y & V \\ \hline
A  &  0 &  19 &  25 &  16 &  50 &  18 &  21 &  78 &  23 &  26 &  22 &  22 &  19 &  14 &  39 &  28 &  15 &  42 &  35 &  8     \\
R &  8 &  0 &  15 &  23 &  23 &  11 &  15 &  37 &  12 &  12 &  19 &  18 &  9 &  15 &  23 &  18 &  11 &  49 &  18 &  9     \\
N & 15 &  33 &  0 &  32 &  51 &  19 &  18 &  56 &  16 &  19 &  11 &  19 &  50 &  21 &  33 &  21 &  16 &  42 &  22 &  14    \\
D  &  7 &  29 &  22 &  0 &  57 &  21 &  25 &  42 &  11 &  19 &  23 &  19 &  16 &  21 &  44 &  16 &  15 &  32 &  16 &  11    \\
C & 2 &1 &1 &2 &0 &7 &1 &9 &8 &4 &5 &5 &2 &2 &2 &4 &0 &8 &2 &5     \\
Q & 5 &12 &18 &12 &33 &0 &14 &42 &15 &18 &7 &15 &21 &7 &33 &5 &7 &32 &21 &18    \\
E &  14 &  14 &  14 &  50 &  47 &  30 &  0 &  62 &  15 &  35 &  19 &  19 &  18 &  29 &  32 &  16 &  11 &  79 &  19 &  11    \\
G &  18 &  11 &  19 &  19 &  62 &  18 &  18 &  0 &  23 &  26 &  29 &  9 &  18 &  25 &  29 &  22 &  12 &  46 &  15 &  9   \\
H &  0 &  15 &  4 &  15 &  11 &  14 &  9 &  28 &  0 &  5 &  9 &  7 &  5 &  12 &  9 &  9 &  2 &  21 &  9 &  7    \\
I &  9 &  16 &  16 &  14 &  46 &  16 &  25 &  58 &  33 &  0 &  49 &  19 &  56 &  35 &  29 &  14 &  14 &  54 &  18 &  32   \\
L &  5 &  28 &  23 &  50 &  57 &  30 &  25 &  70 &  21 &  53 &  0 &  19 &  47 &  40 &  40 &  21 &  19 &  51 &  22 &  33    \\
K &  2 &  44 &  22 &  56 &  78 &  22 &  28 &  51 &  22 &  35 &  18 &  0 &  29 &  19 &  47 &  7 &  11 &  49 &  21 &  15   \\
M &  4 &  5 &  5 &  5 &  9 &  8 &  8 &  26 &  5 &  12 &  28 &  5 &  0 &  7 &  9 &  5 &  4 &  16 &  7 &  8     \\
F&  8 &  19 &  16 &  25 &  35 &  18 &  9 &  36 &  14 &  21 &  19 &  7 &  21 &  0 &  21 &  9 &  5 &  46 &  21 &  2     \\
P &  8 &  11 &  5 &  12 &  16 &  9 &  14 &  25 &  9 &  28 &  9 &  14 &  28 &  16 &  0 &  4 &  14 &  30 &  16 &  2     \\
S &  21 &  26 &  22 &  33 &  49 &  14 &  28 &  53 &  23 &  30 &  26 &  21 &  29 &  22 &  36 &  0 &  40 &  65 &  36 &  19   \\
T &  9 &  19 &  21 &  35 &  33 &  22 &  21 &  40 &  9 &  33 &  42 &  22 &  30 &  28 &  29 &  22 &  0 &  47 &  25 &  19     \\
W &  0 &  2 &  4 &  4 &  7 &  1 &  1 &  12 &  4 &  7 &  5 &  0 &  5 &  7 &  7 &  4 &  0 &  0 &  7 &  4     \\
Y&  7 &  9 &  16 &  23 &  25 &  18 &  15 &  36 &  11 &  9 &  5 &  7 &  29 &  26 &  15 &  16 &  8 &  37 &  0 &  9    \\
V &  8 &  12 &  7 &  29 &  44 &  25 &  9 &  54 &  25 &  49 &  14 &  14 &  43 &  19 &  11 &  15 &  11 &  40 &  18 &  0 
\end{tabular}
% \end{adjustbox}
\label{tab:conservative_statis}
\vspace{-5pt}
\end{table*}

\begin{table*}[h]
\caption{Structural Radical Statistics}
\centering
\vspace{-5pt}
% \begin{adjustbox}{width=.7\linewidth}
\begin{tabular}{l|llllllllllllllllllll}
  & A  & R  & N  & D  & C  & Q & E  & G  & H  & I  & L  & K  & M  & F  & P  & S  & T  & W  & Y  & V  \\ \hline
A  &  0 &  28 &  22 &  16 &  39 &  5 &  19 &  22 &  30 &  28 &  25 &  5 &  25 &  22 &  33 &  14 &  14 &  64 &  16 &  14     \\
R &  8 &  0 &  11 &  33 &  19 &  5 &  28 &  22 &  16 &  25 &  8 &  2 &  11 &  36 &  25 &  5 &  5 &  33 &  11 &  11     \\
N &  11 &  16 &  0 &  8 &  47 &  11 &  22 &  16 &  14 &  19 &  25 &  22 &  19 &  25 &  25 &  5 &  8 &  25 &  14 &  19     \\
D  &  11 &  11 &  11 &  0 &  25 &  14 &  22 &  16 &  19 &  25 &  11 &  16 &  44 &  16 &  16 &  11 &  0 &  22 &  14 &  25    \\
C  &  2 &  0 &  0 &  0 &  0 &  2 &  2 &  11 &  2 &  0 &  8 &  8 &  2 &  2 &  5 &  8 &  5 &  2 &  0 &  5     \\
Q &  2 &  19 &  5 &  11 &  25 &  0 &  22 &  16 &  14 &  14 &  14 &  16 &  14 &  16 &  25 &  5 &  5 &  19 &  8 &  8     \\
E &  5 &  11 &  5 &  19 &  58 &  5 &  0 &  22 &  14 &  19 &  14 &  11 &  28 &  36 &  19 &  5 &  8 &  39 &  25 &  19     \\
G &  2 &  25 &  2 &  16 &  56 &  11 &  14 &  0 &  8 &  33 &  28 &  22 &  53 &  28 &  14 &  14 &  5 &  44 &  22 &  8     \\
H &  2 &  5 &  0 &  0 &  16 &  14 &  8 &  11 &  0 &  11 &  8 &  8 &  0 &  0 &  14 &  2 &  0 &  5 &  5 &  14     \\
I  &  25 &  28 &  22 &  36 &  33 &  5 &  22 &  44 &  5 &  0 &  2 &  8 &  16 &  8 &  30 &  11 &  11 &  47 &  5 &  11     \\
L  &  22 &  28 &  25 &  30 &  64 &  28 &  28 &  72 &  19 &  25 &  0 &  47 &  33 &  11 &  56 &  28 &  22 &  36 &  28 &  19     \\
K  &  14 &  2 &  8 &  25 &  81 &  22 &  19 &  30 &  11 &  14 &  8 &  0 &  16 &  30 &  33 &  5 &  19 &  64 &  19 &  19      \\
M  &  2 &  0 &  16 &  11 &  5 &  2 &  2 &  25 &  8 &  8 &  0 &  8 &  0 &  2 &  19 &  2 &  0 &  14 &  5 &  5  \\
F &  16 &  11 &  11 &  22 &  28 &  8 &  28 &  19 &  11 &  16 &  11 &  19 &  14 &  0 &  39 &  19 &  2 &  11 &  8 &  5      \\
P  &  8 &  11 &  2 &  19 &  22 &  5 &  16 &  11 &  8 &  2 &  14 &  16 &  36 &  14 &  0 &  2 &  5 &  36 &  22 &  8  \\
S  &  22 &  8 &  5 &  14 &  44 &  16 &  22 &  30 &  22 &  28 &  28 &  28 &  25 &  22 &  25 &  0 &  16 &  58 &  16 &  14  \\
T  &  11 &  14 &  16 &  22 &  56 &  8 &  19 &  42 &  14 &  5 &  19 &  8 &  33 &  22 &  19 &  14 &  0 &  58 &  16 &  11   \\
W  &  2 &  0 &  0 &  2 &  2 &  8 &  2 &  14 &  2 &  2 &  0 &  0 &  2 &  2 &  5 &  5 &  2 &  0 &  2 &  8   \\
Y &  25 &  14 &  2 &  5 &  28 &  5 &  8 &  25 &  16 &  11 &  5 &  19 &  25 &  8 &  19 &  8 &  8 &  14 &  0 &  11  \\
V&  8 &  19 &  16 &  36 &  25 &  19 &  30 &  53 &  14 &  8 &  11 &  11 &  44 &  16 &  19 &  11 &  8 &  56 &  8 &  0  
\end{tabular}
% \end{adjustbox}
\label{tab:radical_statis}
\vspace{-5pt}
\end{table*}

%%%%%%%%%%%%%%%%%%%%%%%%%%%%%%%%%%%%%%%%%%%%%%%%%%%%%%%%%%%%%%%%%%%%%%%%%%%%%%%
%%%%%%%%%%%%%%%%%%%%%%%%%%%%%%%%%%%%%%%%%%%%%%%%%%%%%%%%%%%%%%%%%%%%%%%%%%%%%%%

\bibliographystyle{plainnat}
\bibliography{main}

\begin{thebibliography}{66}
\providecommand{\natexlab}[1]{#1}
\providecommand{\url}[1]{\texttt{#1}}
\expandafter\ifx\csname urlstyle\endcsname\relax
  \providecommand{\doi}[1]{doi: #1}\else
  \providecommand{\doi}{doi: \begingroup \urlstyle{rm}\Url}\fi

\bibitem[Ackers and Smith(1985)]{ackers1985effects}
Gary~K Ackers and Francine~R Smith.
\newblock Effects of site-specific amino acid modification on protein
  interactions and biological function.
\newblock \emph{Annual review of biochemistry}, 54\penalty0 (1):\penalty0
  597--629, 1985.

\bibitem[Ahdritz et~al.(2022)Ahdritz, Bouatta, Kadyan, Xia, Gerecke,
  O{\textquoteright}Donnell, Berenberg, Fisk, Zanichelli, Zhang, Nowaczynski,
  Wang, Stepniewska-Dziubinska, Zhang, Ojewole, Guney, Biderman, Watkins, Ra,
  Lorenzo, Nivon, Weitzner, Ban, Sorger, Mostaque, Zhang, Bonneau, and
  AlQuraishi]{Ahdritz2022.11.20.517210}
Gustaf Ahdritz, Nazim Bouatta, Sachin Kadyan, Qinghui Xia, William Gerecke,
  Timothy~J O{\textquoteright}Donnell, Daniel Berenberg, Ian Fisk, Niccolò
  Zanichelli, Bo~Zhang, Arkadiusz Nowaczynski, Bei Wang, Marta~M
  Stepniewska-Dziubinska, Shang Zhang, Adegoke Ojewole, Murat~Efe Guney, Stella
  Biderman, Andrew~M Watkins, Stephen Ra, Pablo~Ribalta Lorenzo, Lucas Nivon,
  Brian Weitzner, Yih-En~Andrew Ban, Peter~K Sorger, Emad Mostaque, Zhao Zhang,
  Richard Bonneau, and Mohammed AlQuraishi.
\newblock Openfold: Retraining alphafold2 yields new insights into its learning
  mechanisms and capacity for generalization.
\newblock \emph{bioRxiv}, 2022.
\newblock \doi{10.1101/2022.11.20.517210}.

\bibitem[AlQuraishi(2021)]{alquraishi2021machine}
Mohammed AlQuraishi.
\newblock Machine learning in protein structure prediction.
\newblock \emph{Current opinion in chemical biology}, 65:\penalty0 1--8, 2021.

\bibitem[Arpino et~al.(2014)Arpino, Reddington, Halliwell, Rizkallah, and
  Jones]{arpino2014random}
James~AJ Arpino, Samuel~C Reddington, Lisa~M Halliwell, Pierre~J Rizkallah, and
  D~Dafydd Jones.
\newblock Random single amino acid deletion sampling unveils structural
  tolerance and the benefits of helical registry shift on gfp folding and
  structure.
\newblock \emph{Structure}, 22\penalty0 (6):\penalty0 889--898, 2014.

\bibitem[Betts and Russell(2003)]{betts2003amino}
Matthew~J Betts and Robert~B Russell.
\newblock Amino acid properties and consequences of substitutions.
\newblock \emph{Bioinformatics for geneticists}, 317:\penalty0 289, 2003.

\bibitem[Blumer et~al.(1987)Blumer, Ehrenfeucht, Haussler, and
  Warmuth]{blumer1987occam}
Anselm Blumer, Andrzej Ehrenfeucht, David Haussler, and Manfred~K Warmuth.
\newblock Occam's razor.
\newblock \emph{Information processing letters}, 24\penalty0 (6):\penalty0
  377--380, 1987.

\bibitem[Bordo and Argos(1991)]{bordo1991suggestions}
Domenico Bordo and Patrick Argos.
\newblock Suggestions for “safe” residue substitutions in site-directed
  mutagenesis.
\newblock \emph{Journal of molecular biology}, 217\penalty0 (4):\penalty0
  721--729, 1991.

\bibitem[Candes and Tao(2005)]{candes2005decoding}
Emmanuel~J Candes and Terence Tao.
\newblock Decoding by linear programming.
\newblock \emph{IEEE transactions on information theory}, 51\penalty0
  (12):\penalty0 4203--4215, 2005.

\bibitem[Carter(1986)]{carter1986site}
Paul Carter.
\newblock Site-directed mutagenesis.
\newblock \emph{Biochemical Journal}, 237\penalty0 (1):\penalty0 1, 1986.

\bibitem[Chen et~al.(2016)Chen, Xu, Zhang, and Guestrin]{chen2016training}
Tianqi Chen, Bing Xu, Chiyuan Zhang, and Carlos Guestrin.
\newblock Training deep nets with sublinear memory cost.
\newblock \emph{arXiv preprint arXiv:1604.06174}, 2016.

\bibitem[Choi and Chan(2015)]{choi2015provean}
Yongwook Choi and Agnes~P Chan.
\newblock Provean web server: a tool to predict the functional effect of amino
  acid substitutions and indels.
\newblock \emph{Bioinformatics}, 31\penalty0 (16):\penalty0 2745--2747, 2015.

\bibitem[Cito et~al.(2022)Cito, Dillig, Murali, and
  Chandra]{cito2022counterfactual}
J{\"u}rgen Cito, Isil Dillig, Vijayaraghavan Murali, and Satish Chandra.
\newblock Counterfactual explanations for models of code.
\newblock In \emph{Proceedings of the 44th International Conference on Software
  Engineering: Software Engineering in Practice}, pages 125--134, 2022.

\bibitem[Clemmons(2001)]{clemmons2001use}
David~R Clemmons.
\newblock Use of mutagenesis to probe igf-binding protein structure/function
  relationships.
\newblock \emph{Endocrine reviews}, 22\penalty0 (6):\penalty0 800--817, 2001.

\bibitem[Cohen et~al.(2009)Cohen, Huang, Chen, Benesty, Benesty, Chen, Huang,
  and Cohen]{cohen2009pearson}
Israel Cohen, Yiteng Huang, Jingdong Chen, Jacob Benesty, Jacob Benesty,
  Jingdong Chen, Yiteng Huang, and Israel Cohen.
\newblock Pearson correlation coefficient.
\newblock \emph{Noise reduction in speech processing}, pages 1--4, 2009.

\bibitem[Dagan et~al.(2002)Dagan, Talmor, and Graur]{dagan2002ratios}
Tal Dagan, Yael Talmor, and Dan Graur.
\newblock Ratios of radical to conservative amino acid replacement are affected
  by mutational and compositional factors and may not be indicative of positive
  darwinian selection.
\newblock \emph{Molecular biology and evolution}, 19\penalty0 (7):\penalty0
  1022--1025, 2002.

\bibitem[Dill and MacCallum(2012)]{dill2012protein}
Ken~A Dill and Justin~L MacCallum.
\newblock The protein-folding problem, 50 years on.
\newblock \emph{science}, 338\penalty0 (6110):\penalty0 1042--1046, 2012.

\bibitem[Dill et~al.(2008)Dill, Ozkan, Shell, and Weikl]{dill2008protein}
Ken~A Dill, S~Banu Ozkan, M~Scott Shell, and Thomas~R Weikl.
\newblock The protein folding problem.
\newblock \emph{Annual review of biophysics}, 37:\penalty0 289, 2008.

\bibitem[Doering et~al.(2018)Doering, Lee, Kristiansen, Evenseth, Barron,
  Sylte, and LaLone]{doering2018silico}
Jon~A Doering, Sehan Lee, Kurt Kristiansen, Linn Evenseth, Mace~G Barron,
  Ingebrigt Sylte, and Carlie~A LaLone.
\newblock In silico site-directed mutagenesis informs species-specific
  predictions of chemical susceptibility derived from the sequence alignment to
  predict across species susceptibility (seqapass) tool.
\newblock \emph{Toxicological Sciences}, 166\penalty0 (1):\penalty0 131--145,
  2018.

\bibitem[Dominy and Andrews(2003)]{dominy2003site}
Clifford~N Dominy and David~W Andrews.
\newblock Site-directed mutagenesis by inverse pcr.
\newblock In \emph{E. coli Plasmid Vectors}, pages 209--223. Springer, 2003.

\bibitem[Edgar and Batzoglou(2006)]{edgar2006multiple}
Robert~C Edgar and Serafim Batzoglou.
\newblock Multiple sequence alignment.
\newblock \emph{Current opinion in structural biology}, 16\penalty0
  (3):\penalty0 368--373, 2006.

\bibitem[Egli et~al.(2006)Egli, Flavell, Pyle, Wilson, Haq, Luisi, Fisher,
  Laughton, Allen, and Engels]{10.1039/9781847555380}
Martin Egli, Andy Flavell, Anna~Marie Pyle, W~David Wilson, S~Ihtshamul Haq,
  Ben Luisi, Julie Fisher, Charlie Laughton, Stephanie Allen, and Joachim
  Engels.
\newblock \emph{Chapter 5.6 Nucleic Acids in Biotechnology}.
\newblock The Royal Society of Chemistry, 2006.
\newblock ISBN 978-0-85404-654-6.
\newblock \doi{10.1039/9781847555380}.

\bibitem[Epstein(1967)]{epstein1967non}
Charles~J Epstein.
\newblock Non-randomness of ammo-acid changes in the evolution of homologous
  proteins.
\newblock \emph{Nature}, 215\penalty0 (5099):\penalty0 355--359, 1967.

\bibitem[Flores-Ram{\'\i}rez et~al.(2007)Flores-Ram{\'\i}rez, Rivera,
  Morales-Pablos, Osuna, Sober{\'o}n, and Gayt{\'a}n]{flores2007effect}
Gabriela Flores-Ram{\'\i}rez, Manuel Rivera, Alfredo Morales-Pablos, Joel
  Osuna, Xavier Sober{\'o}n, and Paul Gayt{\'a}n.
\newblock The effect of amino acid deletions and substitutions in the longest
  loop of gfp.
\newblock \emph{BMC chemical biology}, 7\penalty0 (1):\penalty0 1--10, 2007.

\bibitem[Gl{\"u}ck and Wool(2002)]{gluck2002analysis}
Anton Gl{\"u}ck and Ira~G Wool.
\newblock Analysis by systematic deletion of amino acids of the action of the
  ribotoxin restrictocin.
\newblock \emph{Biochimica et Biophysica Acta (BBA)-Protein Structure and
  Molecular Enzymology}, 1594\penalty0 (1):\penalty0 115--126, 2002.

\bibitem[Glymour et~al.(2016)Glymour, Pearl, and Jewell]{glymour2016causal}
Madelyn Glymour, Judea Pearl, and Nicholas~P Jewell.
\newblock \emph{Causal inference in statistics: A primer}.
\newblock John Wiley \& Sons, 2016.

\bibitem[Goyal et~al.(2019)Goyal, Wu, Ernst, Batra, Parikh, and
  Lee]{goyal2019counterfactual}
Yash Goyal, Ziyan Wu, Jan Ernst, Dhruv Batra, Devi Parikh, and Stefan Lee.
\newblock Counterfactual visual explanations.
\newblock In \emph{International Conference on Machine Learning}, pages
  2376--2384. PMLR, 2019.

\bibitem[Guidotti et~al.(2019)Guidotti, Monreale, Giannotti, Pedreschi,
  Ruggieri, and Turini]{guidotti2019factual}
Riccardo Guidotti, Anna Monreale, Fosca Giannotti, Dino Pedreschi, Salvatore
  Ruggieri, and Franco Turini.
\newblock Factual and counterfactual explanations for black box decision
  making.
\newblock \emph{IEEE Intelligent Systems}, 34\penalty0 (6):\penalty0 14--23,
  2019.

\bibitem[Guo et~al.(2004)Guo, Choe, and Loeb]{guo2004protein}
Haiwei~H Guo, Juno Choe, and Lawrence~A Loeb.
\newblock Protein tolerance to random amino acid change.
\newblock \emph{Proceedings of the National Academy of Sciences}, 101\penalty0
  (25):\penalty0 9205--9210, 2004.

\bibitem[Hutchison et~al.(1978)Hutchison, Phillips, Edgell, Gillam, Jahnke, and
  Smith]{hutchison1978mutagenesis}
Clyde~A Hutchison, Sandra Phillips, Marshall~H Edgell, Shirley Gillam, Patricia
  Jahnke, and Michael Smith.
\newblock Mutagenesis at a specific position in a dna sequence.
\newblock \emph{Journal of Biological Chemistry}, 253\penalty0 (18):\penalty0
  6551--6560, 1978.

\bibitem[Ilari and Savino(2008)]{ilari2008protein}
Andrea Ilari and Carmelinda Savino.
\newblock Protein structure determination by x-ray crystallography.
\newblock \emph{Bioinformatics}, pages 63--87, 2008.

\bibitem[Jumper et~al.(2021)Jumper, Evans, Pritzel, Green, Figurnov,
  Ronneberger, Tunyasuvunakool, Bates, {\v{Z}}{\'\i}dek, Potapenko,
  et~al.]{jumper2021highly}
John Jumper, Richard Evans, Alexander Pritzel, Tim Green, Michael Figurnov,
  Olaf Ronneberger, Kathryn Tunyasuvunakool, Russ Bates, Augustin
  {\v{Z}}{\'\i}dek, Anna Potapenko, et~al.
\newblock Highly accurate protein structure prediction with alphafold.
\newblock \emph{Nature}, 596\penalty0 (7873):\penalty0 583--589, 2021.

\bibitem[Kato et~al.(2018)Kato, Piel, Reid, Gaston, Ohene-Frempong,
  Krishnamurti, Smith, Panepinto, Weatherall, Costa, et~al.]{kato2018sickle}
Gregory~J Kato, Fr{\'e}d{\'e}ric~B Piel, Clarice~D Reid, Marilyn~H Gaston,
  Kwaku Ohene-Frempong, Lakshmanan Krishnamurti, Wally~R Smith, Julie~A
  Panepinto, David~J Weatherall, Fernando~F Costa, et~al.
\newblock Sickle cell disease.
\newblock \emph{Nature Reviews Disease Primers}, 4\penalty0 (1):\penalty0
  1--22, 2018.

\bibitem[Kingma and Ba(2014)]{kingma2014adam}
Diederik~P Kingma and Jimmy Ba.
\newblock Adam: A method for stochastic optimization.
\newblock \emph{arXiv preprint arXiv:1412.6980}, 2014.

\bibitem[Lampridis et~al.(2020)Lampridis, Guidotti, and
  Ruggieri]{lampridis2020explaining}
Orestis Lampridis, Riccardo Guidotti, and Salvatore Ruggieri.
\newblock Explaining sentiment classification with synthetic exemplars and
  counter-exemplars.
\newblock In \emph{International Conference on Discovery Science}, pages
  357--373. Springer, 2020.

\bibitem[Li et~al.(2022)Li, Ji, and Zhang]{li2022from}
Zelong Li, Jianchao Ji, and Yongfeng Zhang.
\newblock {From Kepler to Newton: Explainable {AI} for Science Discovery}.
\newblock In \emph{ICML 2022 2nd AI for Science Workshop}, 2022.

\bibitem[Lin et~al.(2021)Lin, Lan, and Li]{lin2021generative}
Wanyu Lin, Hao Lan, and Baochun Li.
\newblock Generative causal explanations for graph neural networks.
\newblock In \emph{International Conference on Machine Learning}, pages
  6666--6679. PMLR, 2021.

\bibitem[Maas et~al.(2013)Maas, Hannun, Ng, et~al.]{maas2013rectifier}
Andrew~L Maas, Awni~Y Hannun, Andrew~Y Ng, et~al.
\newblock Rectifier nonlinearities improve neural network acoustic models.
\newblock In \emph{Proc. icml}, volume~30, page~3. Atlanta, Georgia, USA, 2013.

\bibitem[Mart{\'\i}nez et~al.(2020)Mart{\'\i}nez, H{\"u}ttelmaier, and
  Bertoldo]{martinez2020unveiling}
Dail{\'e}n~G Mart{\'\i}nez, Stefan H{\"u}ttelmaier, and Jean~B Bertoldo.
\newblock Unveiling druggable pockets by site-specific protein modification:
  Beyond antibody-drug conjugates.
\newblock \emph{Frontiers in Chemistry}, 8:\penalty0 586942, 2020.

\bibitem[Masso and Vaisman(2008)]{masso2008accurate}
Majid Masso and Iosif~I Vaisman.
\newblock Accurate prediction of stability changes in protein mutants by
  combining machine learning with structure based computational mutagenesis.
\newblock \emph{Bioinformatics}, 24\penalty0 (18):\penalty0 2002--2009, 2008.

\bibitem[Masso et~al.(2006)Masso, Lu, and Vaisman]{masso2006computational}
Majid Masso, Zhibin Lu, and Iosif~I Vaisman.
\newblock Computational mutagenesis studies of protein structure-function
  correlations.
\newblock \emph{Proteins: Structure, Function, and Bioinformatics}, 64\penalty0
  (1):\penalty0 234--245, 2006.

\bibitem[Miyata et~al.(1979)Miyata, Miyazawa, and Yasunaga]{miyata1979two}
Takashi Miyata, Sanzo Miyazawa, and Teruo Yasunaga.
\newblock Two types of amino acid substitutions in protein evolution.
\newblock \emph{Journal of molecular evolution}, 12:\penalty0 219--236, 1979.

\bibitem[Motohashi(2015)]{motohashi2015simple}
Ken Motohashi.
\newblock A simple and efficient seamless dna cloning method using slice from
  escherichia coli laboratory strains and its application to slip site-directed
  mutagenesis.
\newblock \emph{BMC biotechnology}, 15\penalty0 (1):\penalty0 1--9, 2015.

\bibitem[Moult et~al.()Moult, Fidelis, Kryshtafovych, Schwede, and
  Topf]{moult14critical}
J~Moult, K~Fidelis, A~Kryshtafovych, T~Schwede, and M~Topf.
\newblock Critical assessment of techniques for protein structure prediction,
  fourteenth round.
\newblock \emph{CASP 14 Abstract Book}.

\bibitem[Sarkar and Sommer(1990)]{sarkar1990megaprimer}
Gobinda Sarkar and Steve~S Sommer.
\newblock The" megaprimer" method of site-directed mutagenesis.
\newblock \emph{Biotechniques}, 8\penalty0 (4):\penalty0 404--407, 1990.

\bibitem[Senior et~al.(2020)Senior, Evans, Jumper, Kirkpatrick, Sifre, Green,
  Qin, {\v{Z}}{\'\i}dek, Nelson, Bridgland, et~al.]{senior2020improved}
Andrew~W Senior, Richard Evans, John Jumper, James Kirkpatrick, Laurent Sifre,
  Tim Green, Chongli Qin, Augustin {\v{Z}}{\'\i}dek, Alexander~WR Nelson, Alex
  Bridgland, et~al.
\newblock Improved protein structure prediction using potentials from deep
  learning.
\newblock \emph{Nature}, 577\penalty0 (7792):\penalty0 706--710, 2020.

\bibitem[Sotomayor-Vivas et~al.(2022)Sotomayor-Vivas, Hern{\'a}ndez-Lemus, and
  Dorantes-Gilardi]{sotomayor2022linking}
Cristina Sotomayor-Vivas, Enrique Hern{\'a}ndez-Lemus, and Rodrigo
  Dorantes-Gilardi.
\newblock Linking protein structural and functional change to mutation using
  amino acid networks.
\newblock \emph{Plos one}, 17\penalty0 (1):\penalty0 e0261829, 2022.

\bibitem[Spicer and Davis(2014)]{spicer2014selective}
Christopher~D Spicer and Benjamin~G Davis.
\newblock Selective chemical protein modification.
\newblock \emph{Nature communications}, 5\penalty0 (1):\penalty0 1--14, 2014.

\bibitem[Stenson et~al.(2017)Stenson, Mort, Ball, Evans, Hayden, Heywood,
  Hussain, Phillips, and Cooper]{stenson2017human}
Peter~D Stenson, Matthew Mort, Edward~V Ball, Katy Evans, Matthew Hayden, Sally
  Heywood, Michelle Hussain, Andrew~D Phillips, and David~N Cooper.
\newblock The human gene mutation database: towards a comprehensive repository
  of inherited mutation data for medical research, genetic diagnosis and
  next-generation sequencing studies.
\newblock \emph{Human genetics}, 136:\penalty0 665--677, 2017.

\bibitem[Studer et~al.(2013)Studer, Dessailly, and Orengo]{studer2013residue}
Romain~A Studer, Benoit~H Dessailly, and Christine~A Orengo.
\newblock Residue mutations and their impact on protein structure and function:
  detecting beneficial and pathogenic changes.
\newblock \emph{Biochemical journal}, 449\penalty0 (3):\penalty0 581--594,
  2013.

\bibitem[Szymkowski(2005)]{szymkowski2005creating}
David~E Szymkowski.
\newblock Creating the next generation of protein therapeutics through rational
  drug design.
\newblock \emph{CURRENT OPINION IN DRUG DISCOVERY AND DEVELOPMENT}, 8\penalty0
  (5):\penalty0 590, 2005.

\bibitem[Tan et~al.(2021)Tan, Xu, Ge, Li, Chen, and
  Zhang]{tan2021counterfactual}
Juntao Tan, Shuyuan Xu, Yingqiang Ge, Yunqi Li, Xu~Chen, and Yongfeng Zhang.
\newblock Counterfactual explainable recommendation.
\newblock In \emph{Proceedings of the 30th ACM International Conference on
  Information \& Knowledge Management}, pages 1784--1793, 2021.

\bibitem[Tan et~al.(2022)Tan, Geng, Fu, Ge, Xu, Li, and Zhang]{tan2022learning}
Juntao Tan, Shijie Geng, Zuohui Fu, Yingqiang Ge, Shuyuan Xu, Yunqi Li, and
  Yongfeng Zhang.
\newblock Learning and evaluating graph neural network explanations based on
  counterfactual and factual reasoning.
\newblock In \emph{Proceedings of the ACM Web Conference 2022}, pages
  1018--1027, 2022.

\bibitem[Tan et~al.(2020)Tan, Wu, Wei, and Li]{tan2020chemical}
Yi~Tan, Hongxiang Wu, Tongyao Wei, and Xuechen Li.
\newblock Chemical protein synthesis: advances, challenges, and outlooks.
\newblock \emph{Journal of the American Chemical Society}, 142\penalty0
  (48):\penalty0 20288--20298, 2020.

\bibitem[Tolkachev et~al.(2022)Tolkachev, Mell, Zdancewic, and
  Bastani]{tolkachev2022counterfactual}
George Tolkachev, Stephen Mell, Stephan Zdancewic, and Osbert Bastani.
\newblock Counterfactual explanations for natural language interfaces.
\newblock In \emph{Proceedings of the 60th Annual Meeting of the Association
  for Computational Linguistics}, pages 113--118, 2022.

\bibitem[Torrisi et~al.(2020)Torrisi, Pollastri, and Le]{torrisi2020deep}
Mirko Torrisi, Gianluca Pollastri, and Quan Le.
\newblock Deep learning methods in protein structure prediction.
\newblock \emph{Computational and Structural Biotechnology Journal},
  18:\penalty0 1301--1310, 2020.

\bibitem[Vermeire et~al.(2022)Vermeire, Brughmans, Goethals, de~Oliveira, and
  Martens]{vermeire2022explainable}
Tom Vermeire, Dieter Brughmans, Sofie Goethals, Raphael Mazzine~Barbossa
  de~Oliveira, and David Martens.
\newblock Explainable image classification with evidence counterfactual.
\newblock \emph{Pattern Analysis and Applications}, pages 1--21, 2022.

\bibitem[Wachter et~al.(2017)Wachter, Mittelstadt, and
  Russell]{wachter2017counterfactual}
Sandra Wachter, Brent Mittelstadt, and Chris Russell.
\newblock Counterfactual explanations without opening the black box: Automated
  decisions and the gdpr.
\newblock \emph{Harv. JL \& Tech.}, 31:\penalty0 841, 2017.

\bibitem[Wiltgen(2009)]{wiltgen2009structualbio}
Marco Wiltgen.
\newblock Structural bioinformatics: From the sequence to structure and
  function.
\newblock \emph{Current Bioinformatics}, 4:\penalty0 54--87, 01 2009.
\newblock \doi{10.2174/157489309787158170}.

\bibitem[Xu and Zhang(2010)]{xu2010significant}
Jinrui Xu and Yang Zhang.
\newblock How significant is a protein structure similarity with tm-score= 0.5?
\newblock \emph{Bioinformatics}, 26\penalty0 (7):\penalty0 889--895, 2010.

\bibitem[Yampolsky and Stoltzfus(2005)]{yampolsky2005exchangeability}
Lev~Y Yampolsky and Arlin Stoltzfus.
\newblock The exchangeability of amino acids in proteins.
\newblock \emph{Genetics}, 170\penalty0 (4):\penalty0 1459--1472, 2005.

\bibitem[Yang et~al.(2020)Yang, Kenny, Ng, Yang, Smyth, and
  Dong]{yang2020generating}
Linyi Yang, Eoin~M Kenny, Tin Lok~James Ng, Yi~Yang, Barry Smyth, and Ruihai
  Dong.
\newblock Generating plausible counterfactual explanations for deep
  transformers in financial text classification.
\newblock \emph{arXiv preprint arXiv:2010.12512}, 2020.

\bibitem[Zemla(2003)]{zemla2003lga}
Adam Zemla.
\newblock Lga: a method for finding 3d similarities in protein structures.
\newblock \emph{Nucleic acids research}, 31\penalty0 (13):\penalty0 3370--3374,
  2003.

\bibitem[Zhang(2000)]{zhang2000rates}
Jianzhi Zhang.
\newblock Rates of conservative and radical nonsynonymous nucleotide
  substitutions in mammalian nuclear genes.
\newblock \emph{Journal of molecular evolution}, 50\penalty0 (1):\penalty0
  56--68, 2000.

\bibitem[Zhang et~al.(2018)Zhang, Case, and Peng]{zhang2018propagated}
Meiling Zhang, David~A Case, and Jeffrey~W Peng.
\newblock Propagated perturbations from a peripheral mutation show interactions
  supporting ww domain thermostability.
\newblock \emph{Structure}, 26\penalty0 (11):\penalty0 1474--1485, 2018.

\bibitem[Zhang and Skolnick(2004)]{zhang2004scoring}
Yang Zhang and Jeffrey Skolnick.
\newblock Scoring function for automated assessment of protein structure
  template quality.
\newblock \emph{Proteins: Structure, Function, and Bioinformatics}, 57\penalty0
  (4):\penalty0 702--710, 2004.

\bibitem[Zhang and Skolnick(2005)]{zhang2005tm}
Yang Zhang and Jeffrey Skolnick.
\newblock Tm-align: a protein structure alignment algorithm based on the
  tm-score.
\newblock \emph{Nucleic acids research}, 33\penalty0 (7):\penalty0 2302--2309,
  2005.

\end{thebibliography}

\end{document}